\documentclass{article}

    \usepackage[numbers,sort]{natbib}

\usepackage[preprint]{neurips_2024}

\usepackage[utf8]{inputenc} 
\usepackage[T1]{fontenc}    
\usepackage{hyperref}       
\usepackage{url}            
\usepackage{booktabs}       
\usepackage{amsfonts}       
\usepackage{nicefrac}       
\usepackage{microtype}      
\usepackage{xcolor}         

\usepackage{amsmath}
\DeclareMathOperator*{\argmax}{arg\,max}
\DeclareMathOperator*{\argmin}{arg\,min}

\usepackage{bbm}
\usepackage{dsfont}
\usepackage{graphicx}

\usepackage{color}

\usepackage{algpseudocode}
\algtext*{EndWhile}
\algtext*{EndIf}
\algtext*{EndFunction}
\MakeRobust{\Call}
 
 \usepackage{algorithm}
 \usepackage{algorithmicx}

\algrenewcommand\algorithmicrequire{\textbf{Input:}}
\algrenewcommand\algorithmicensure{\textbf{Output:}}

\usepackage{multirow}
\usepackage{booktabs}
\usepackage{wrapfig}
\usepackage{colortbl}

\usepackage{caption} 
\usepackage{array,ragged2e}
\newcolumntype{R}[1]{>{\RaggedLeft\arraybackslash}p{#1}}
\newcolumntype{L}[1]{>{\RaggedRight\arraybackslash}p{#1}}

\title{Optimizing Class-Level Probability Reweighting Coefficients for Equitable Prompting Accuracy}

\author{Ruixi Lin \qquad Yang You\\
  Department of Computer Science\\
  National University of Singapore\\
  \texttt{\{ruixi,youy\}@comp.nus.edu.sg} \\}

\begin{document}

\maketitle

\begin{abstract}
Even as we engineer LLMs for alignment and safety, they often uncover biases from pre-training data's statistical regularities (from disproportionate co-occurrences to stereotypical associations mirroring human cognitive biases). This leads to persistent, uneven class accuracy in classification and QA. Such per-class accuracy disparities are not inherently resolved by architectural/training evolutions or data scaling, making post-hoc correction essential for equitable performance. To mitigate LLM class accuracy imbalance, we develop a post-hoc probability reweighting method that directly optimizes for non-differentiable performance-driven and fairness-aligned metrics, through a novel COBias metric that highlights disparities in class accuracies. This post-hoc bias mitigation method is grounded in discrete optimization with nonlinear integer programming (NIP) objectives and an efficient metaheuristic solution framework with theoretical convergence guarantees. Operating model-agnostically, it learns reweighting coefficients from output class probabilities to adjust LLM inference outputs without internal weight updates. Evaluations demonstrate its effectiveness: reducing COBias (61\% relative reduction), increasing overall accuracy (18\% relative increase), and achieving robust within-task generalization across diverse prompt configurations.
\end{abstract}

\section{Introduction}
\label{sec:intro}


Despite our goal to build aligned and safe AI systems that act in accordance with human values, large language models (LLMs) consistently acquire biases imprinted in the statistical regularities of their pre-training data, from disproportionate co-occurrence frequencies to stereotypical associations mirroring human cognitive biases \citep{Caliskan2017,Bender2021}. This inevitably leads to persistent, imbalanced accuracy across classes in classification and QA. Even with continued model evolution in training and data, LLMs will not inherently resolve these per-class performance disparities. Consequently, ensuring equitable and robust LLM performance across all classes underscore the necessity of post-hoc correction. 

Without manipulating LLM internal weights, post-hoc correction emerges as a pragmatic approach to mitigate class accuracy imbalance. To this end, we introduce a post-hoc probability reweighting method, the \textit{Debiasing as Nonlinear Integer Programming} (\textbf{DNIP}) method, which directly optimizes non-differentiable performance metrics through discrete optimization for fairer prompting accuracy. 

\begin{wrapfigure}[22]{R}{0.4\textwidth}
  \centering
    \includegraphics[width=\linewidth]{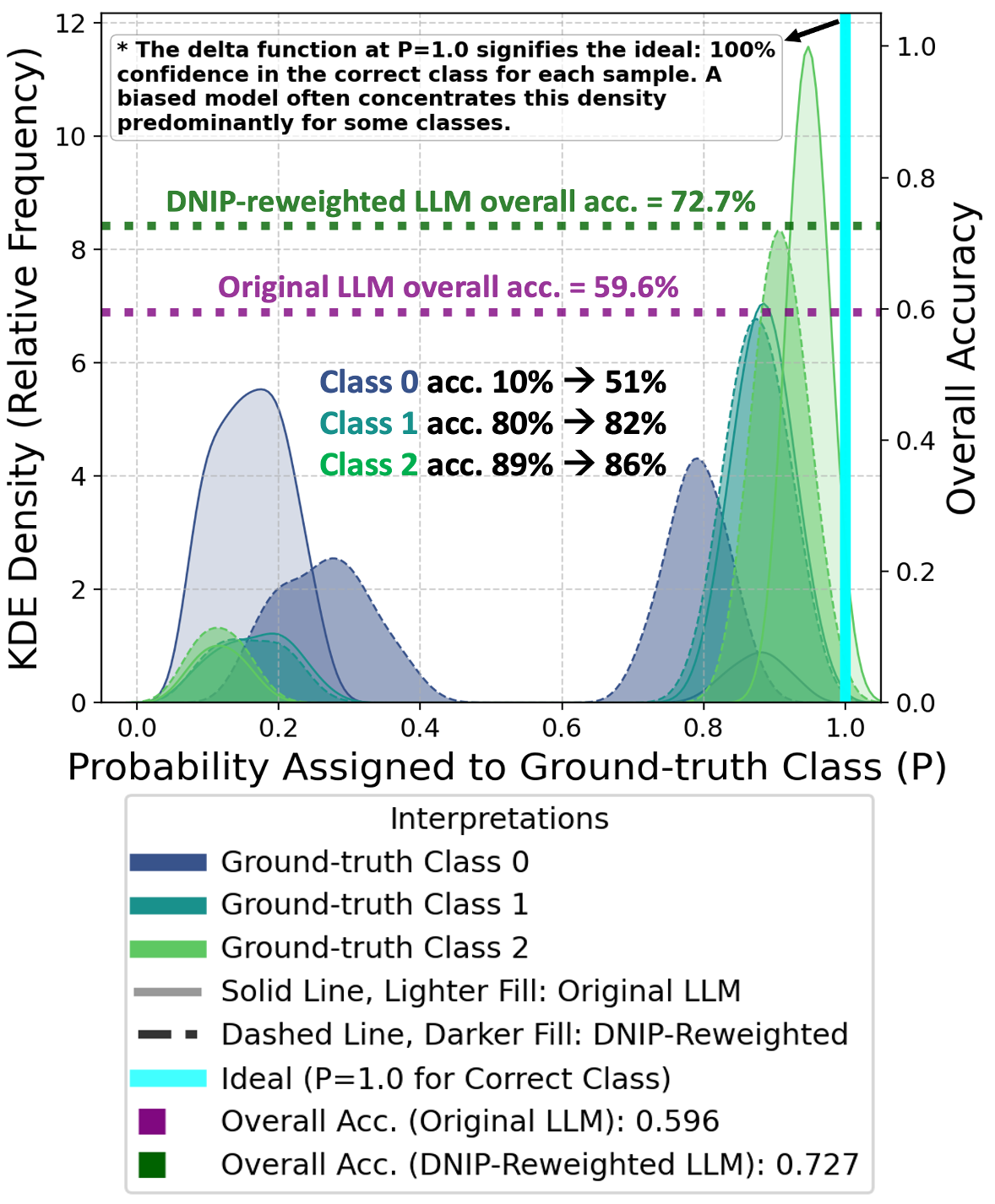}
  \caption{DNIP pushes forward more balanced output probability distribution and fairer class accuracy.}
  \label{fig:kde}
\end{wrapfigure}

DNIP optimizes for a novel \textit{Contextual Oddity Bias} (\textbf{COBias}) metric, which precisely targets class accuracy differences, alongside overall accuracy metrics. It is `trained' on a labeled \textit{optimization} set, where each input sample is represented by the softmax class probability vector obtained at the LLM's answer token (labels are essential for metric calculations). The variables DNIP learns are class-specific reweighting coefficients (represented as discrete weight scale indices), which then directly adjust \textit{test} class probabilities. Crucially, while both the optimization and test sets can be drawn from the same underlying data distribution, the optimization strictly uses samples from the optimization set only, involving no test set data or its derived metric scores, thereby preventing direct test set leakage.

Figure \ref{fig:kde} visualizes how our DNIP correction method effectively mitigates class accuracy imbalances, presenting Kernel Density Estimation (KDE) plots from 5,000 simulated test instances. These plots illustrate the distribution of probabilities an LLM assigns to each sample's \textit{ground-truth class}, revealing inconsistent per-class confidence in original LLM outputs. After DNIP reweighting, they demonstrate a key transformation: reshaping the original uneven distributions into notably more balanced distributions, thereby achieving more uniform class accuracy.

The \textbf{key contributions of DNIP} are enabling: \textbf{1.} Direct modeling of non-differentiable, highly nonlinear performance metrics, such as accuracy and a novel COBias metric. \textbf{2.} Direct use of discrete variables for probability reweighting coefficient selection, not requiring continuous variables.

Under varied LLM prompting strategies, empirical evaluations show that DNIP achieves significant COBias reduction (61\% relative reduction) and accuracy improvement (18\% relative increase), averaged across five LLMs of different scales and families and seven general-domain and biomedical-domain text classification benchmarks.

\section{Related Work}
\label{sec:relatedwork}
Our work aligns with a paradigm shift from indirect, loss-based bias/fairness interventions to direct, outcome-driven optimization for equitable classification. Below, we introduce existing methods and perpetual challenges.\\

\noindent \textbf{Sources of Label Prediction Bias in LLMs, Fairness Metrics.} Label prediction bias in LLMs stems from token imbalances in pretraining data \citep{naviglisurvey} and is further influenced by prompt-specific cues \citep{kassner2020,zhao2021,fuzhao}. Specifically, under-represented tokens can lead LLMs to favor generating common patterns or specific outputs. This multifaceted bias also manifests as sensitivity to prompt templates, demonstrations, and order \citep{jiang2020,holtzman2021,turpin2023}. Regarding fairness, existing metrics evaluate inter-class accuracy discrepancy \citep{benz2020} or deviation from average accuracy \citep{croce2021}. However, while applicable to LLMs, these metrics can lack insight into semantic relations between classes. For instance, approaches aiming for a uniform output distribution might not prevent models from assigning higher probability mass to the preferred classes, causing accuracy imbalances to persist. These layered challenges in bias identification necessitates post-hoc debiasing methods agnostic to prompts and model internals.\\

\noindent \textbf{Bias Mitigation via Retraining.} Classical bias mitigation solutions focus on retraining with loss reweighting or minimizing the maximal loss across all classes. Approaches such as class-balanced losses \citep{cui2019} and focal losses \citep{lin2017focal} reweight loss contributions per class to prevent dominant classes from overwhelming minority ones. Others adopt minimax formulations to optimize worst-case performance across classes \citep{namkoong2016}.

However, loss-based retraining is indirect, and it is sometimes prohibitive for large LLMs due to their immense model size, associated costs, and limited access to closed models. In contrast, our approach is post-hoc without needing to modify LLM parameters, prompt formats, or requiring access to pretraining data.\\

\noindent \textbf{Post-hoc Calibration Methods.} Post-hoc calibration primarily mitigates biases stemming from prompts. These methods utilize content-free \citep{zhao2021} or content-based prompts \citep{bc} to calibrate output class logits/probability distributions and address prompt brittleness. For instance, calibration using a prompt structure without actual content (e.g., ``N/A'') \citep{zhao2021} estimates and subtracts the model's inherent bias towards certain classes, making responses less sensitive to exact prompt wording that might otherwise trigger the bias. Content-based calibration \citep{bc} further mitigates biases emerging from the prompt context, by estimating and subtracting a contextual prior for each class, computed by averaging the output probabilities from a batch of unlabeled, content-based test prompts.

However, prompt-focused interventions are neither explicit nor sufficient in minimizing class accuracy imbalance, while being more uniform in probabilities assigned to ground-truth classes. Some of these classes still receive flatter probability ranges than others and thus lower accuracy, arising from deeper model biases. In Section \ref{sec:exp}, we compare DNIP with content-based calibration, demonstrating the importance of explicitly mitigating class accuracy disparities.\\

\noindent \textbf{Nonlinear Integer Programming and Why It Suits.} Standard gradient-based approaches like logistic regression are unsuited for optimizing non-differentiable metrics. While it is tempting to formulate problems with vector inputs (e.g., class probabilities) and a vector output (e.g., class-level reweighting coefficients) using a metric-surrogate loss, such an approach does \textbf{not directly optimize the actual non-differentiable objectives}; moreover, it can be tedious to form the reweighting coefficient as a continuous scalar for learning, when \textbf{a discrete variable for coefficient selection proves to be sufficiently effective and robust}.

The above challenge is why nonlinear integer programming (NIP) becomes a better fit. NIP enables \textbf{directly modeling non-differentiable, highly nonlinear metrics}, and naturally uses discrete variables (e.g., weight index selectors for coefficients). Therefore, the NIP formulation aligns with our goal to efficiently optimize for our metrics. For backgrounds, integer programming is a paradigm within combinatorial optimization. It formulates constrained discrete optimization problems and find optimal solutions from a finite candidate set \citep{50yearsip}. It is widely applied in NLP, from structured prediction, grammatical error correction, to LLM alignment \citep{martins2009,lin2021,garmendia2024survey}. While standard solvers excel at linear programs, NIP models present challenges due to discontinuities, requiring complex subproblem relaxations. We, however, efficiently solve our model with metaheuristics, specifically simulated annealing, which offers a probabilistic guarantee of convergence to global optima \citep{kirkpatrick1983}.

\section{Quantifying Class Accuracy Imbalance:\\The Contextual Oddity Bias Metric}
\label{sec:cobias}
Building upon the established class accuracy imbalance, often evidenced by disproportionate over-predictions and under-predictions, we introduce the Contextual Oddity Bias (COBias) as a post-hoc bias metric. An optimally debiased LLM would simultaneously enforce minimal COBias while maintaining good overall classification accuracy.

\begin{table}[ht]
    \centering
    \small
    \begin{center}
  \begin{tabular}{R{2cm} R{0.6cm} R{0.6cm}R{0.8cm} R{0.5cm}R{0.4cm}R{0.4cm}}
  \toprule
    \textbf{} &  &  & \textbf{\textit{Pred.}} &  & & \\
    & \textbf{World} & \textbf{Sports} & \textbf{Business} & \textbf{Tech} & \textbf{All} & \textbf{Acc.} \\
    \midrule
    \textbf{World} & 1093 & 64 & 126 & 3 & 1286 & 0.85\\
    \textbf{Sports} & 9 & 1247 & 14 & 0 & 1270& 0.98\\
    \textbf{\textit{True}} \textbf{Business} & 25 & 4 & 1167 & 8 & \textbf{\color{red}1204\color{black}}& \textbf{\color{red}0.97\color{black}} \\
    \textbf{Tech} & 156 & 27 & \textbf{\color{red}822\color{black}} & 235 & \textbf{\color{red}1240\color{black}} & \textbf{\color{red}0.19\color{black}}\\
    \textbf{All} & 1283 & 1342 & \textbf{\color{red}2129\color{black}} & \textbf{\color{red}246\color{black}} & 5000 & -\\
    \textbf{Prec.} & 0.85 & 0.93 & 0.55 & 0.96 & - & -\\
    \bottomrule
    \end{tabular}
    \end{center}
\caption{Over-prediction in class \textit{Business} and under-prediction in class \textit{Tech} for AGNews in Llama-2-13B, shown by the test confusion matrix.}
\label{tab:llama213b_ag}
\end{table}

Class accuracy imbalance in LLMs manifests as over-prediction (a class is predicted more often than its true occurrences) and under-prediction (a class is predicted less often). For instance, as shown in Table \ref{tab:llama213b_ag} for Llama-2-13B on AGNews, \textit{Business} is significantly over-predicted while \textit{Tech} is severely under-predicted, leading to a mere 19\% accuracy for \textit{Tech}. To comprehensively quantify this inter-class disparity, we propose the \textbf{Contextual Oddity Bias (COBias)}, which measures the average absolute difference in accuracy between all pairs of classes:
\begin{equation}
    \text{COBias} = \binom{N}{2}^{-1} \displaystyle \sum_{i=1}^{N-1} \sum_{j=i+1}^{N} \bigg | A_{c_i} - A_{c_j} \bigg |.
    \label{eq:2}
\end{equation}
where $A_{c_i}$ and $A_{c_j}$ denote the classification accuracies for classes $c_i$ and $c_j$ respectively, and $N$ is the total number of classes. This metric directly reflects the overall inter-class accuracy imbalance, providing a robust and well-rounded measurement for scenarios demanding minimal class accuracy disparities. Additionally, the metric name derives from the `odd class' phenomenon of specific, dominant mispredictions. When exclusively capturing biases related to specific dominant misprediction patterns (e.g., \textit{Tech} most-frequently misclassified as \textit{Business}), we also define $\text{COBias}_{\text{single}}$ as the average absolute accuracy difference between each ground-truth class and its most-frequently mispredicted counterpart (details and evaluations in Section \ref{appdix:a} of Technical Appendix).

\textbf{Evaluations consistently reveal significant COBias across various LLMs and NLP tasks.} Using GPT-2-XL \citep{gpt2}, Gemma-2-2B \citep{gemma2}, and Llama-2 (7B, 13B, 70B) \citep{touvron2023llama2openfoundation} across seven diverse classification tasks, we find COBias in all models exceeds 32\% in the 1-shot case, averaging 37.6\% (Table \ref{tab:main} top part (\textit{ICL}) for full results). COBias persists despite increases in model scale or common prompting strategies, with balanced demonstrations only partially mitigating COBias to 25.6\%. This highlights that COBias is a pervasive issue not easily resolved by model scaling or prompting adjustments, necessitating targeted debiasing approaches.

\textbf{Reproducibility}: our \textbf{evaluation tasks} encompass seven diverse classification datasets from news classification to biomedical QA. Five general-domain datasets are: 4-way news topic classification AGNews \citep{zhang2015}, 14-way ontology classification DBpedia \citep{auer2007}, 5-way sentiment classification SST-5 \citep{socher2013}, 6-way retrieval question classification TREC \citep{voorhees2000,li2002}, and binary entailment recognition RTE \citep{dagan2006}. Two biomedical datasets from BLURB \citep{gu2021} are: 5-way biomedical relation extraction DDI \citep{ddi}, and 3-way biomedical question answering PubMedQA \citep{pubmedqa}. Table \ref{tab:main}'s scores use the standard test splits of each dataset, with 5,000 randomly selected test examples for AGNews. For \textbf{LLM classification}, we follow a common practice to predict the $\argmax$ class using probabilities assigned to label tokens, where these probabilities are derived at the LLM's generated answer position. In details, after obtaining softmax probabilities over the vocabulary at answer position, we extract and normalize probabilities corresponding to label tokens existing in the vocabulary to form class probabilities, i.e., $\boldsymbol{p}_m=(p_{m1},\dots, p_{mN})$ for instance $x_m$. The final prediction $\hat{y}_m$ is then $\argmax_{j \in \{1,\dots,N\}} \boldsymbol{p}_{mj}$. To investigate the role of \textbf{prompting strategies} in bias mitigation, all models are initially evaluated under 1-shot ICL settings. For Llama-2-13B, we additionally explore 5-shot and $N$-shot class-balanced demonstrations (where one example from each of $N$ classes is randomly selected). To account for demonstration variance, we conduct three ICL runs per model and dataset using different random seeds to select 1-/5-shot demonstrations, reporting average test scores. For \textbf{hardware}, prompting is performed on an 80G A100 GPU.


\begin{table*}[ht]
\LARGE
\centering
\resizebox{\textwidth}{!}{
 \begingroup
 \renewcommand{\arraystretch}{1.5}
\begin{tabular}{@{}lcccccccccccccccc@{}}
\toprule
\multicolumn{1}{l|}{\multirow[t]{3}{*}{\Huge Model}} & \multicolumn{8}{c|}{\Huge Accuracy} & \multicolumn{8}{c}{\Huge COBias} \\ \cmidrule(l){2-17} 
\multicolumn{1}{l|}{} & \multirow{2}{*}{All} & \multicolumn{5}{c}{General Domain} & \multicolumn{2}{c|}{Biomedical Domain} & \multirow{2}{*}{All} & \multicolumn{5}{c}{General Domain} & \multicolumn{2}{c}{Biomedical Domain} \\ \cmidrule(lr){3-9} \cmidrule(l){11-17} 
\multicolumn{1}{l|}{} &  & AGN & DBP & SST & TREC & RTE & DDI & \multicolumn{1}{c|}{PMQA} &  & AGN & DBP & SST & TREC & RTE & DDI & PMQA \\ \midrule
\multicolumn{17}{c}{\textbf{\textit{\Huge ICL}}} \\ \midrule
\multicolumn{1}{l|}{\Huge GPT-2-XL, 1-shot} & \cellcolor[HTML]{DEEFFC} \textbf{\Huge 38.8} & {\Huge 52.1}\textsubscript{5.4} & {\Huge 31.8}\textsubscript{9.9} & {\Huge 34.9}\textsubscript{13.7} & {\Huge 27.4}\textsubscript{10.5} & {\Huge 55.4}\textsubscript{1.9} & {\Huge 14.5}\textsubscript{4.4} & \multicolumn{1}{c|}{{\Huge 55.2}\textsubscript{0}} & \cellcolor[HTML]{DEEFFC} \textbf{\Huge 50.3} & {\Huge 35.5}\textsubscript{11.5} & {\Huge 40.0}\textsubscript{3.6} & {\Huge 48.7}\textsubscript{5.4} & {\Huge 45.6}\textsubscript{8.7} & {\Huge 82.4}\textsubscript{24.5} & {\Huge 40.7}\textsubscript{5.9} & {\Huge 59.4}\textsubscript{12.6} \\
\multicolumn{1}{l|}{\Huge Gemma-2-2B, 1-shot} & \cellcolor[HTML]{DEEFFC} \textbf{\Huge 53.4} & {\Huge 79.9}\textsubscript{3.4} & {\Huge 72.6}\textsubscript{2.7} & {\Huge 27.6}\textsubscript{4.3} & {\Huge 44.6}\textsubscript{4.7} & {\Huge 70.3}\textsubscript{6.6} & {\Huge 15.4}\textsubscript{10.6} & \multicolumn{1}{c|}{{\Huge 63.4}\textsubscript{1.0}} & \cellcolor[HTML]{DEEFFC} \textbf{\Huge 38.5} & {\Huge 25.7}\textsubscript{6.9} & {\Huge 36.6}\textsubscript{1.8} & {\Huge 42.0}\textsubscript{1.5} & {\Huge 50.2}\textsubscript{0.7} & {\Huge 25.2}\textsubscript{13.1} & {\Huge 35.1}\textsubscript{13.0} & {\Huge 54.4}\textsubscript{3.2} \\
\multicolumn{1}{l|}{\Huge Llama-2-7B, 1-shot} & \cellcolor[HTML]{DEEFFC} \textbf{\Huge 56.8} & {\Huge 86.4}\textsubscript{2.5} & {\Huge 88.9}\textsubscript{2.0} & {\Huge 42.11}\textsubscript{1.1} & {\Huge 66.7}\textsubscript{6.6} & {\Huge 66.3}\textsubscript{4.3} & {\Huge 6.7}\textsubscript{0.4} & \multicolumn{1}{c|}{{\Huge 40.3}\textsubscript{6.7}} & \cellcolor[HTML]{DEEFFC} \textbf{\Huge 37.2} & {\Huge 14.0}\textsubscript{6.5} & {\Huge 13.5}\textsubscript{2.1} & {\Huge 55.6}\textsubscript{1.5} & {\Huge 33.2}\textsubscript{10.0} & {\Huge 61.6}\textsubscript{10.5} & {\Huge 41.4}\textsubscript{1.7} & {\Huge 40.9}\textsubscript{16.1} \\
\multicolumn{1}{l|}{\Huge Llama-2-13B, 1-shot} & \cellcolor[HTML]{DEEFFC} \textbf{\Huge 59.4} & {\Huge 79.9}\textsubscript{7.0} & {\Huge 88.6}\textsubscript{1.7} & {\Huge 44.9}\textsubscript{4.3} & {\Huge 68.5}\textsubscript{10.8} & {\Huge 71.5}\textsubscript{2.2} & {\Huge 7.2}\textsubscript{0.9} & \multicolumn{1}{c|}{{\Huge 55.1}\textsubscript{2.9}} & \cellcolor[HTML]{DEEFFC} \textbf{\Huge 40.5} & {\Huge 28.3}\textsubscript{16.1} & {\Huge 16.2}\textsubscript{3.7} & {\Huge 53.1}\textsubscript{5.0} & {\Huge 35.9}\textsubscript{6.5} & {\Huge 43.4}\textsubscript{7.0} & {\Huge 45.6}\textsubscript{5.9} & {\Huge 61.2}\textsubscript{1.9} \\
\multicolumn{1}{l|}{\Huge Llama-2-70B, 1-shot*} & \cellcolor[HTML]{DEEFFC} \textbf{\Huge 66.4} & {\Huge 87.4}\textsubscript{2.2} & {\Huge 94.1}\textsubscript{1.0} & {\Huge 41.8}\textsubscript{8.6} & {\Huge 79.3}\textsubscript{2.5} & {\Huge 78.6}\textsubscript{2.6} & {\Huge 8.8}\textsubscript{3.7} & \multicolumn{1}{c|}{{\Huge 74.5}\textsubscript{2.3}} & \cellcolor[HTML]{DEEFFC} \textbf{\Huge 31.8} & {\Huge 14.5}\textsubscript{5.6} & {\Huge 8.3}\textsubscript{1.2} & {\Huge 54.7}\textsubscript{1.4} & {\Huge 22.4}\textsubscript{2.0} & {\Huge 25.5}\textsubscript{7.6} & {\Huge 36.0}\textsubscript{2.1} & {\Huge 61.5}\textsubscript{3.3} \\
\multicolumn{1}{l|}{\Huge Llama-2-13B, 5-shot} & \cellcolor[HTML]{DEEFFC} \textbf{\Huge 63.9} & {\Huge 82.5}\textsubscript{2.0} & {\Huge 93.6}\textsubscript{1.3} & {\Huge 45.8}\textsubscript{6.0} & {\Huge 58.7}\textsubscript{23.3} & {\Huge 61.9}\textsubscript{16.9} & {\Huge 34.1}\textsubscript{42.1} & \multicolumn{1}{c|}{{\Huge 70.4}\textsubscript{7.1}} & \cellcolor[HTML]{DEEFFC} \textbf{\Huge 39.2} & {\Huge 24.4}\textsubscript{4.2} & {\Huge 9.0}\textsubscript{2.0} & {\Huge 48.0}\textsubscript{15.4} & {\Huge 35.4}\textsubscript{13.7} & {\Huge 61.3}\textsubscript{40.4} & {\Huge 44.4}\textsubscript{5.4} & {\Huge 52.2}\textsubscript{19.3} \\
\multicolumn{1}{l|}{\Huge Llama-2-13B, N-shot} & \cellcolor[HTML]{DEEFFC} \textbf{\Huge 61.9} & {\Huge 83.5}\textsubscript{1.5} & {\Huge 95.2}\textsubscript{1.2} & {\Huge 50.3}\textsubscript{2.3} & {\Huge 67.0}\textsubscript{12.7} & {\Huge 75.0}\textsubscript{0.8} & {\Huge 9.7}\textsubscript{1.0} & \multicolumn{1}{c|}{{\Huge 52.3}\textsubscript{5.3}} & \cellcolor[HTML]{DEEFFC} \textbf{\Huge 25.6} & {\Huge 14.9}\textsubscript{5.1} & {\Huge 7.0}\textsubscript{2.2} & {\Huge 36.3}\textsubscript{7.2} & {\Huge 38.2}\textsubscript{5.1} & {\Huge 22.5}\textsubscript{13.2} & {\Huge 39.7}\textsubscript{3.5} & {\Huge 20.9}\textsubscript{4.2} \\ \midrule

\multicolumn{17}{c}{\textbf{\textit{\Huge DNIP}}} \\ \midrule\multicolumn{1}{l|}{\Huge GPT-2-XL, 1-shot} & \cellcolor[HTML]{E5D0FF} \textbf{\Huge 54.4}\textsubscript{1.0} & \cellcolor[HTML]{F1DDFF} {\Huge 68.5}\textsubscript{1.0} & \cellcolor[HTML]{E5D0FF} {\Huge 69.9}\textsubscript{9.1} & \cellcolor[HTML]{FAD6FF} {\Huge 44.5}\textsubscript{2.20} & \cellcolor[HTML]{FEE8EC} {\Huge 46.3}\textsubscript{12.7} & \cellcolor[HTML]{FFFFFF} {\Huge 50.8}\textsubscript{2.1} & \cellcolor[HTML]{FEE8EC} {\Huge 43.9}\textsubscript{14.9} & \multicolumn{1}{c|}{\cellcolor[HTML]{FFF2F5} {\Huge 57.1}\textsubscript{1.3}} & \cellcolor[HTML]{E5D0FF} \textbf{\Huge 18.9}\textsubscript{2.4} & \cellcolor[HTML]{E5D0FF} {\Huge 1.4}\textsubscript{0.5} & \cellcolor[HTML]{FAD6FF} {\Huge 24.1}\textsubscript{8.3} & \cellcolor[HTML]{FAD6FF} {\Huge 26.0}\textsubscript{2.5} & \cellcolor[HTML]{FAD6FF} {\Huge 27.2}\textsubscript{7.2} & \cellcolor[HTML]{F1DDFF} {\Huge 7.1}\textsubscript{5.0} & \cellcolor[HTML]{FFEBEE} {\Huge 17.0}\textsubscript{7.1} & \multicolumn{1}{c}{\cellcolor[HTML]{FAD6FF} {\Huge 29.8}\textsubscript{25.0}} \\
\multicolumn{1}{l|}{\Huge Gemma-2-2B, 1-shot} & \cellcolor[HTML]{FFEBEE} \textbf{\Huge 65.1}\textsubscript{1.1} & \cellcolor[HTML]{FAD6FF} {\Huge 86.9}\textsubscript{1.1} & \cellcolor[HTML]{FAD6FF} {\Huge 88.4}\textsubscript{0.9} & \cellcolor[HTML]{FFEBEE} {\Huge 33.2}\textsubscript{7.9} & \cellcolor[HTML]{E5D0FF} {\Huge 69.6}\textsubscript{0.8} & \cellcolor[HTML]{FFF2F5} {\Huge 72.6}\textsubscript{2.3} & \cellcolor[HTML]{FFEBEE} {\Huge 41.5}\textsubscript{18.2\%} & \multicolumn{1}{c|}{\cellcolor[HTML]{FFFFFF} {\Huge 63.4}\textsubscript{11.3}} & \cellcolor[HTML]{FFEBEE} \textbf{\Huge 19.9}\textsubscript{0.4} & \cellcolor[HTML]{FAD6FF} {\Huge 7.1}\textsubscript{0.5} & \cellcolor[HTML]{E5D0FF} {\Huge 14.0}\textsubscript{1.0} & \cellcolor[HTML]{FFF2F5} {\Huge 34.3}\textsubscript{9.9} & \cellcolor[HTML]{F1DDFF} {\Huge 21.4}\textsubscript{2.8} & \cellcolor[HTML]{FFF2F5} {\Huge 2.6}\textsubscript{1.8} & \cellcolor[HTML]{FFF2F5} {\Huge 18.3}\textsubscript{7.2} & \multicolumn{1}{c}{\cellcolor[HTML]{FFEBEE} {\Huge 41.8}\textsubscript{27.3}} \\
\multicolumn{1}{l|}{\Huge Llama-2-7B, 1-shot} & \cellcolor[HTML]{FFEBEE} \textbf{\Huge 68.5}\textsubscript{0.4} & \cellcolor[HTML]{FFF2F5} {\Huge 86.7}\textsubscript{0.4} & \cellcolor[HTML]{FFF2F5} {\Huge 92.9}\textsubscript{0.4} & \cellcolor[HTML]{E5D0FF} {\Huge 50.6}\textsubscript{2.7} & \cellcolor[HTML]{FFF2F5} {\Huge 68.1}\textsubscript{1.0} & \cellcolor[HTML]{FAD6FF} {\Huge 73.9}\textsubscript{2.3} & \cellcolor[HTML]{FFEBEE} {\Huge 44.5}\textsubscript{3.8} & \multicolumn{1}{c|}{\cellcolor[HTML]{F1DDFF} {\Huge 62.7}\textsubscript{8.3}} & \cellcolor[HTML]{FAD6FF} \textbf{\Huge 14.5}\textsubscript{1.1} & \cellcolor[HTML]{FFEBEE} {\Huge 1.3}\textsubscript{0.1} & \cellcolor[HTML]{FFF2F5} {\Huge 7.7}\textsubscript{0.6} & \cellcolor[HTML]{FAD6FF} {\Huge 28.0}\textsubscript{21.6} & \cellcolor[HTML]{FFEBEE} {\Huge 15.9}\textsubscript{1.6} & \cellcolor[HTML]{FAD6FF} {\Huge 1.9}\textsubscript{1.8} & \cellcolor[HTML]{FFEBEE} {\Huge 11.6}\textsubscript{3.0} & \multicolumn{1}{c}{\cellcolor[HTML]{FFF2F5} {\Huge 35.4}\textsubscript{22.8}} \\
\multicolumn{1}{l|}{\Huge Llama-2-13B, 1-shot} & \cellcolor[HTML]{FFEBEE} \textbf{\Huge 69.2}\textsubscript{0.7} & \cellcolor[HTML]{FAD6FF} {\Huge 87.9}\textsubscript{0.7} & \cellcolor[HTML]{FFF2F5} {\Huge 93.4}\textsubscript{0.6} & \cellcolor[HTML]{FAD6FF} {\Huge 48.3}\textsubscript{1.9} & \cellcolor[HTML]{FAD6FF} {\Huge 77.1}\textsubscript{2.0} & \cellcolor[HTML]{FFF2F5} {\Huge 74.3}\textsubscript{0.8} & \cellcolor[HTML]{FFEBEE} {\Huge 40.4}\textsubscript{6.0} & \multicolumn{1}{c|}{\cellcolor[HTML]{FAD6FF} {\Huge 63.1}\textsubscript{14.0}} & \cellcolor[HTML]{FAD6FF} \textbf{\Huge 14.3}\textsubscript{1.0} & \cellcolor[HTML]{FFEBEE} {\Huge 6.3}\textsubscript{0.6} & \cellcolor[HTML]{FFF2F5} {\Huge 7.7}\textsubscript{0.6} & \cellcolor[HTML]{E5D0FF} {\Huge 18.7}\textsubscript{10.1} & \cellcolor[HTML]{FAD6FF} {\Huge 14.2}\textsubscript{1.3} & \cellcolor[HTML]{FAD6FF} {\Huge 4.3}\textsubscript{3.3} & \cellcolor[HTML]{E5D0FF} {\Huge 7.5}\textsubscript{3.2} & \multicolumn{1}{c}{\cellcolor[HTML]{FAD6FF} {\Huge 41.1}\textsubscript{29.6}} \\
\multicolumn{1}{l|}{\Huge Llama-2-70B, 1-shot*} & \cellcolor[HTML]{FFF2F5} \textbf{\Huge 73.6}\textsubscript{0.2} & \cellcolor[HTML]{FFF2F5} {\Huge 89.8}\textsubscript{0.2} & \cellcolor[HTML]{FFF2F5} {\Huge 96.5}\textsubscript{0.5} & \cellcolor[HTML]{FAD6FF} {\Huge 49.4}\textsubscript{2.0} & \cellcolor[HTML]{FFFFFF} {\Huge 79.3}\textsubscript{4.0} & \cellcolor[HTML]{FFF2F5} {\Huge 80.0}\textsubscript{1.1} & \cellcolor[HTML]{E5D0FF} {\Huge 52.5}\textsubscript{14.4} & \multicolumn{1}{c|}{\cellcolor[HTML]{FFFFFF} {\Huge 67.5}\textsubscript{2.4}} & \cellcolor[HTML]{FFF2F5} \textbf{\Huge 17.3}\textsubscript{1.1} & \cellcolor[HTML]{FFF2F5} {\Huge 6.5}\textsubscript{0.4} & \cellcolor[HTML]{FFF2F5} {\Huge 3.8}\textsubscript{0.3} & \cellcolor[HTML]{FFF2F5} {\Huge 47.1}\textsubscript{14.0} & \cellcolor[HTML]{FFF2F5} {\Huge 17.3}\textsubscript{4.8} & \cellcolor[HTML]{FFF2F5} {\Huge 4.6}\textsubscript{1.8} & \cellcolor[HTML]{FFF2F5} {\Huge 16.4}\textsubscript{9.0} & \multicolumn{1}{c}{\cellcolor[HTML]{FAD6FF} {\Huge 25.7}\textsubscript{2.0}} \\
\multicolumn{1}{l|}{\Huge Llama-2-13B, 5-shot} & \cellcolor[HTML]{FFF2F5} \textbf{\Huge 71.9}\textsubscript{0.6} & \cellcolor[HTML]{FAD6FF} {\Huge 88.5}\textsubscript{0.6} & \cellcolor[HTML]{FFF2F5} {\Huge 95.8}\textsubscript{0.7} & \cellcolor[HTML]{FAD6FF} {\Huge 52.9}\textsubscript{2.5} & \cellcolor[HTML]{FAD6FF} {\Huge 76.6}\textsubscript{8.7} & \cellcolor[HTML]{F1DDFF} {\Huge 75.8}\textsubscript{4.5} & \cellcolor[HTML]{FFF2F5} {\Huge 54.1}\textsubscript{5.1} & \multicolumn{1}{c|}{\cellcolor[HTML]{FFFFFF} {\Huge 59.9}\textsubscript{1.4}} & \cellcolor[HTML]{F1DDFF} \textbf{\Huge 10.3}\textsubscript{0.7} & \cellcolor[HTML]{FAD6FF} {\Huge 7.0}\textsubscript{0.9} & \cellcolor[HTML]{FFF2F5} {\Huge 5.7}\textsubscript{1.1} & \cellcolor[HTML]{F1DDFF} {\Huge 15.8}\textsubscript{12.6} & \cellcolor[HTML]{FAD6FF} {\Huge 14.4}\textsubscript{7.5} & \cellcolor[HTML]{FAD6FF} {\Huge 3.1}\textsubscript{1.9} & \cellcolor[HTML]{FFEBEE} {\Huge 16.5}\textsubscript{7.6} & \multicolumn{1}{c}{\cellcolor[HTML]{F1DDFF} {\Huge 9.6}\textsubscript{4.6}} \\
\multicolumn{1}{l|}{\Huge Llama-2-13B, N-shot} & \cellcolor[HTML]{FFEBEE} \textbf{\Huge 70.7}\textsubscript{0.5} & \cellcolor[HTML]{FFEBEE} {\Huge 88.7}\textsubscript{0.5} & \cellcolor[HTML]{FFF2F5} {\Huge 96.6}\textsubscript{0.5} & \cellcolor[HTML]{FFF2F5} {\Huge 51.3}\textsubscript{1.0} & \cellcolor[HTML]{FFEBEE} {\Huge 82.7}\textsubscript{1.4} & \cellcolor[HTML]{FFF2F5} {\Huge 76.7}\textsubscript{3.7} & \cellcolor[HTML]{FFEBEE} {\Huge 43.6}\textsubscript{6.1} & \multicolumn{1}{c|}{\cellcolor[HTML]{FFF2F5} {\Huge 55.3}\textsubscript{2.6}} & \cellcolor[HTML]{FFEBEE} \textbf{\Huge 7.5}\textsubscript{0.6} & \cellcolor[HTML]{FFF2F5} {\Huge 7.3}\textsubscript{0.5} & \cellcolor[HTML]{FFF2F5} {\Huge 4.3}\textsubscript{0.7} & \cellcolor[HTML]{F1DDFF} {\Huge 2.8}\textsubscript{1.5} & \cellcolor[HTML]{F1DDFF} {\Huge 12.1}\textsubscript{5.5} & \cellcolor[HTML]{FFF2F5} {\Huge 5.0}\textsubscript{3.3} & \cellcolor[HTML]{FAD6FF} {\Huge 12.5}\textsubscript{4.3} & \multicolumn{1}{c}{\cellcolor[HTML]{FFEBEE} {\Huge 8.7}\textsubscript{1.2}} \\ \bottomrule
\end{tabular}
\endgroup
}
\caption{\textit{\textbf{ICL}}: Test accuracy and COBias (\%) under ICL settings. 1- or 5-shot: 1 or 5 demonstrations (randomly selected); N-shot: balanced demonstrations from each class. $*$: using reported results in \citep{dcs}. \textit{\textbf{DNIP}}: Results after DNIP reweighting. Pink shaded regions indicate DNIP outperforms original LLM; \textbf{darker} shades, \textbf{larger improvements}.}
\label{tab:main}
\end{table*}

\section{Debiasing as Nonlinear Integer Programming}
\label{sec:method}
Our \textbf{Debiasing as Nonlinear Integer Programming (DNIP)} method works to post-hoc reweight biased LLM class probabilities for fairer prompting accuracy across classes, specifically minimizing COBias while maintaining overall accuracy. Analogous to fine-tuning, DNIP features an optimization stage on a labeled dataset. However, rather than updating LLM parameters or learning continuous coefficients as fine-tuning typically does, DNIP uniquely optimizes a mathematical model to learn discrete, integer-valued weight index selectors. These selectors then map to class-specific reweighting coefficients from a predefined weight scale. The resulting coefficients are then used for adjusting actual test class probabilities. Our experiments confirm that using these discrete variables is effective and robust. 

\subsection{The DNIP Objective}
A key aspect of our method is the direct use of discrete variables for coefficient selection. Formally, we introduce a $K$-point correction weight scale for reweighting coefficient selection, i.e., $\boldsymbol{\omega} = (\omega_1,\dots,\omega_{K})$, with weight values $\omega_k = k/K$. For example, a $10$-point scale has 10 possible weight values to choose from, i.e., $\boldsymbol{\omega} = (0.1, \dots, 1.0)$. Given the predefined weight scale, we define integer variables $\xi=(\xi_{1},\dots,\xi_{N})$ that selects a weight index $k$ for the $n$-th output class from the weight scale. Therefore, $\xi_{n}= k$ maps to the outcome that the $n$-th class selects $\omega_k$ as its reweighting coefficient.

The optimization process searches for the best selection of weight indices for \textit{all} classes. The input to optimization is a labeled optimization set consisting of each instance's output class probabilities and ground-truth label. (The optimization and test process is illustrated in Section \ref{appdix:b} of Technical Appendix for intuitive visualizations.) The DNIP model is:
\begin{align}
  \label{eq:3}
\min \quad & Z(\xi)=\frac{1}{M} \displaystyle \sum_{m=1}^{M} \mathds{1}\{\hat{y}_m \ne y_m\} +  \beta{N\choose 2}^{-1} \sum_{i=1}^{N-1} \sum_{j=i+1}^{N} \bigg | A_{c_i} - A_{c_j} \bigg | - \tau \displaystyle \sum_{j=1}^{N} \textrm{PMI}_j \\
\textrm{s.t.} \quad & \hat{y}_m = \argmax_{n \in \{1,\dots,N\}}  \omega_{\xi_{n}}p_{mn},  \qquad m=1,\dots,M \\
\quad & \textrm{PMI}_j = \textrm{PMI}(\hat{S}^{j}, S^j) = \log \frac{f(\hat{S}^{j}, S^j)}{f(\hat{S}^{j})f(S^j)}, \nonumber \\ \qquad & j=1,\dots,N \\
  & \xi_{n} \in \{1,\dots, K\},  \qquad n=1,\dots,N 
\end{align}
where $y_m$ is the label for instance $x_m$, $\beta$ and $\tau$ are parameters that control the bias and the PMI terms respectively. The PMI term is introduced to maximize the pointwise mutual information between ground-truth class $j$ instances and predicted class $j$, which can be seen as a confidence level, or a constraint to penalize each individual class. $\hat{S}^{j}$ and $S^j$ denote the instances of prediction $j$ and true label $j$ respectively, $f(\hat{S}^{j})$ is the ratio between the number of instances with prediction $j$ and the total number of instances, similarly, $f({S}^j)$ is the ratio between the number of instances labeled $j$ and the total number of instances, $f(\hat{S}^{j}, S^j)$ is the ratio between the number of correct predictions of class $j$ and the total number of instances. In actual computations, we apply add-$\mu$ smoothing to the respective ratios in the numerator and denominator to smooth out zeros, and it is selected along other parameters on the development set. 

Intuitively, the first term directly minimizes error rates, the second term COBias directly minimize class accuracy differences, and the PMI term, is employed to further encourage predictions within a class to be close to the actual class. The reweighted class probabilities and prediction are:
\begin{align}
    \boldsymbol{p}^{\ast}_m &= (\omega_{\xi^{\ast}_{1}}p_{m1}, \dots, \omega_{\xi^{\ast}_{N}}p_{mN}) \\
    \hat{y}^{\ast}_m &= \argmax_{n \in \{1,\dots,N\}}\boldsymbol{p}^{\ast}_m
\end{align} 

\subsection{Solving DNIP with Simulated Annealing}
\setlength\intextsep{0pt}
\begin{wrapfigure}[18]{R}{0.6\textwidth}
    \begin{minipage}{0.6\textwidth}
      \begin{algorithm}[H]
    \caption{Optimizing the selection of correction weights with simulated annealing}
    \label{alg:sa}
    \begin{algorithmic}[1]
        \Require{$(\xi, \boldsymbol{p}, y, \omega)$\textrm{: weight selection variable, class probabilities and labels, the weight scale}}
        \Ensure{$(\xi, \{y^{\ast}_m\}_{1}^M)$}
        \State $T\gets T_{max}$, $\xi^{\ast} \gets \xi\gets$ \textbf{\Call{\color{blue}INIT}{$ $}}
        \While{$T \ge T_{min}$}
            \While{\text{\textit{inner-loop criterion} is not satisfied}}
            \State $\xi_{new}\gets$ \textbf{\Call{\color{blue}PERTURB}{$\xi$}}\Comment{Sample a new $\xi$}
            \State $\Delta z\gets$ $z(\xi_{new})- z(\xi)$
            \If{$\Delta z \le 0$}
                \State $\xi \gets$ $\xi_{new}$
                    \If{$z(\xi_{new}) < z(\xi^{\ast})$}
                    \State $\xi^{\ast} \gets$ $\xi_{new}$
                    \EndIf
            \ElsIf{\Call{random}{$0,1$} $<$ \Call{exp}{$\Delta z/T$}}
            \State $\xi \gets$ $\xi_{new}$ \Comment{Accept a worse $\xi$}
	      \EndIf
            \EndWhile
            \State $T\gets$ $\alpha T$
        \EndWhile
        \State $y^{\ast} \gets$ \textbf{\Call{\color{blue}INFER}{$\xi^{\ast}, \boldsymbol{p}, \omega$}}
        \State \textbf{return} $\xi^{\ast}, y^{\ast}$
    \end{algorithmic}
      \end{algorithm}
    \end{minipage}
  \end{wrapfigure}

Since the DNIP objective function is nonlinear and non-differentiable, finding the optimal weight indices requires a robust search algorithm. We employ Simulated Annealing (SA), a heuristic for solving complex, non-differentiable optimization problems \citep{kirkpatrick1983,eglese1990}. Algorithm \ref{alg:sa} shows SA steps. Notably, based on function evaluation and probabilistic acceptance, SA effectively explores the solution space. The computational complexity is $O(NK)$, and SA is guaranteed to probabilistically converge to global optima (a derivation and proof in Section \ref{appdix:c} of Technical Appendix) and remains practically efficient on our evaluation benchmarks.

\section{Experiments on DNIP}
\label{sec:exp}
\textbf{Experimental Setup.} For the optimization sets, we use labels and output class probabilities of samples obtained from the standard training splits of the same evaluation datasets used for original LLM measurements (despite 10,000 examples for large datasets like AGNews, DBpedia, and DDI). Test sets also comply. Hyperparameter tuning of $\beta, \tau, K, \mu$ is done on the development set. DNIP runs on optimization sets on CPU, and then we reweight test class probabilities. Notably, using the same metrics for both optimization and evaluation is analogous to training a supervised model directly with a metric like Mean Squared Error (MSE) and then evaluating with that same metric.

\subsection{Main Results}
\label{subsec:mainresults}
Results are shown under \textit{DNIP} in Table \ref{tab:main}. COBias was over 32\% before debiasing. By DNIP, it is consistently below 20\% for all models under the 1-shot setting, demonstrating a significantly reduced class accuracy imbalance. Overall, DNIP achieves an average COBias of \textbf{14.7\%} (a \textbf{61\% relative reduction}) and an average accuracy of \textbf{67.6\%} (an \textbf{18\% relative increase}) across all models and tasks.

This improvement is consistent across model sizes. For example, Llama-2-13B sees a 10\% accuracy gain and a 65\% relative COBias reduction (40.5\% $\rightarrow$ 14.3\%), while Llama-2-70B's accuracy improves by 7\% with a 46\% relative COBias reduction (31.8\% $\rightarrow$ 17.3\%). Furthermore, DNIP consistently lowers COBias with improved accuracy under the 5- and k-shot settings, down to 7.5\% in the k-shot case (25.6\% $\rightarrow$ 7.5\%). More findings using smaller optimization sets are shown in Section \ref{appdix:d} of Technical Appendix.

\subsection{Computational Time}
\begin{figure*}[ht]
    \centering
    \begin{minipage}[t]{0.45\textwidth}
        \centering
        \small
        \begin{tabular}{@{}ccc@{}}
            \toprule
            Dataset & Time/s & Variable Space ($NK$) \\
            \midrule
            AGNews & 2278 & 120 \\
            DBpedia & 2584 & 420 \\
            SST-5 & 1499 & 150 \\
            TREC & 1274 & 180 \\
            RTE & 518 & 60 \\
            DDI & 2092 & 150 \\
            PubMedQA & 203 & 90 \\
            \bottomrule
        \end{tabular}
        \captionof{table}{Optimization time for each dataset.}
        \label{tab:opt_time}
    \end{minipage}%
    \hfill
    \begin{minipage}[t]{0.48\textwidth}
        \centering
        \includegraphics[width=\linewidth]{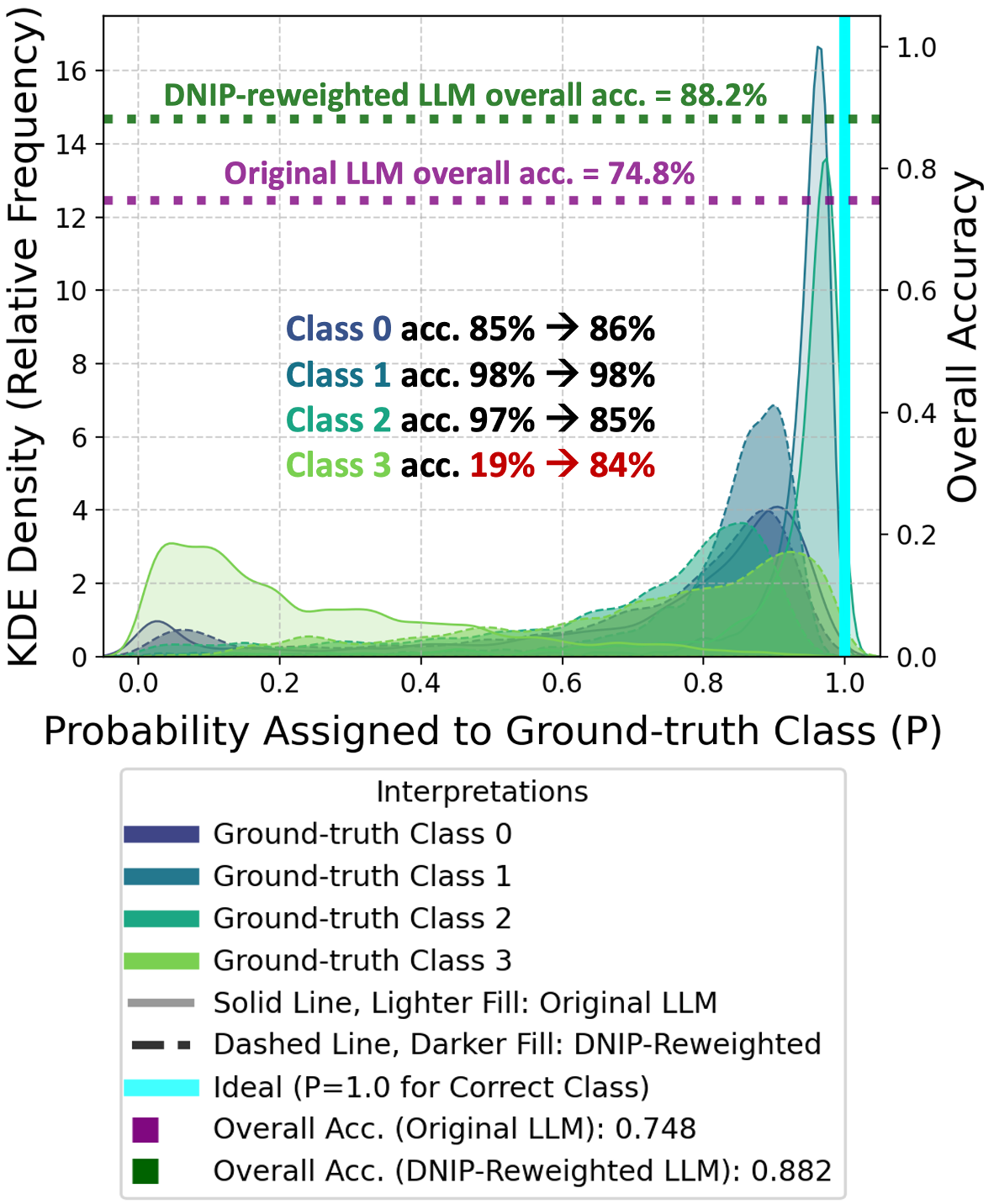}
        \caption{Significantly more balanced output probability distribution and fairer class accuracy on AGNews; using seed0 ICL outputs, DNIP's learned weight indices are $[3, 1, 1, 20]$ for class 0 to 3 respectively, using a 30-point weight scale.}
        \label{fig:kde_ag}
    \end{minipage}
\end{figure*}

DNIP is a practical post-hoc method. For all datasets, optimization \textbf{finishes in minutes}, with the learned weights yielding reusable correction coefficients with minimal overhead (element-wise multiplication and argmax operations) for test inference. A 30-point weight scale (fixing $\beta=2.7, \tau=0.2$) is sufficient for most evaluated datasets, with optimization time shown in Table \ref{tab:opt_time}.

\subsection{KDE Plots}
Using AGNews as an example, we show its KDE plots in Figure \ref{fig:kde_ag}, demonstrating \textbf{significantly more balanced} output probability assigned to each ground-truth class and fairer class accuracy, especially a boost in accuracy in class 3/\textit{Tech} from 19\% to 84\%.
\vspace{5pt}

\subsection{Comparisons to Prior Methods}
\label{subsec:prior}
\begin{figure}[ht]
  \begin{center}
\includegraphics[width=0.8\linewidth]{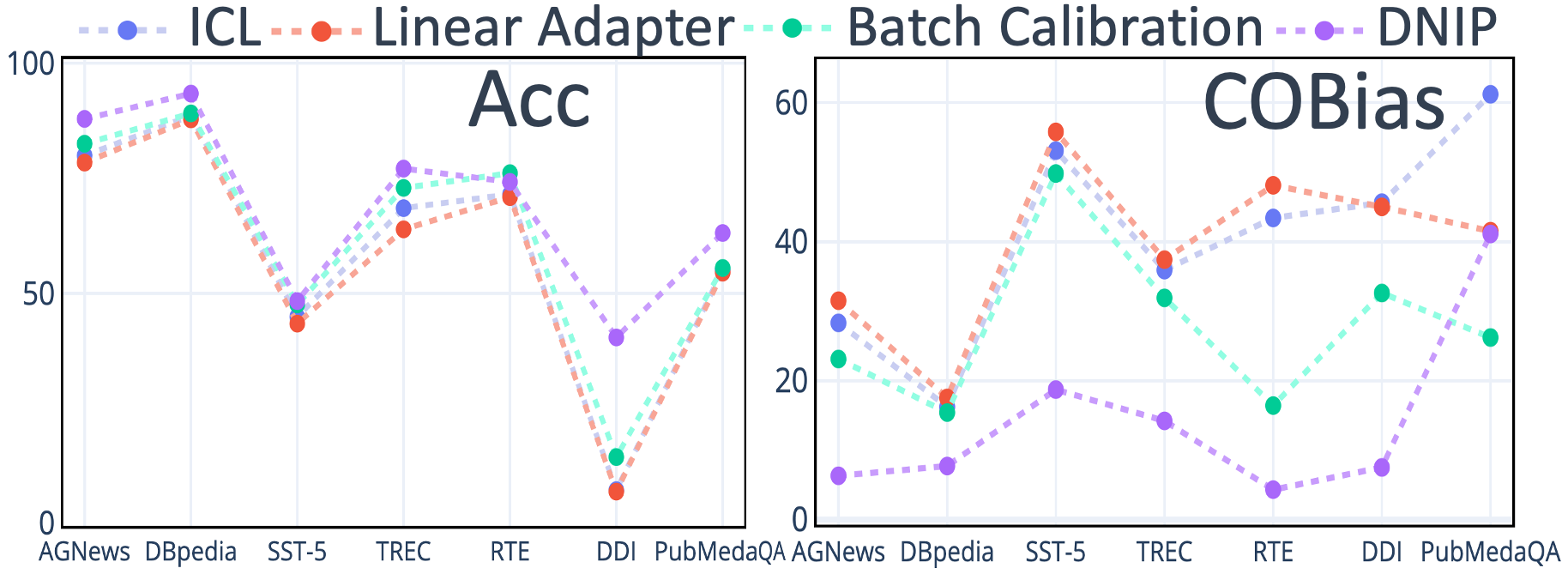}
  \end{center}
  \caption{Comparisons to prior methods.}
  \label{fig:moremethods}
\end{figure}

As introduced in Section \ref{sec:relatedwork}, prior methods tackle label prediction bias from retraining or prompt-based calibration. Therefore, we compare with two popular categories of debiasing methods, adaptations and calibrations. In particular, with the same class probabilities as input, we build a single linear adapter (LA) network with adapter dimension of 64, and train it towards the same objective as DNIP; we also follow the batch calibration method \citep{bc} to estimate a calibration term using test examples. Figure \ref{fig:moremethods} compares the methods. Fine-tuning a simple linear adapter does not work well on the class output probabilities. \textbf{DNIP outperforms both adaptation and calibration baselines in accuracy and COBias.}

\vspace{8pt}

\subsection{Ablations}
\label{subsec:accablations}
\begin{figure*}[ht]
  \centering
  \includegraphics[width=\linewidth]{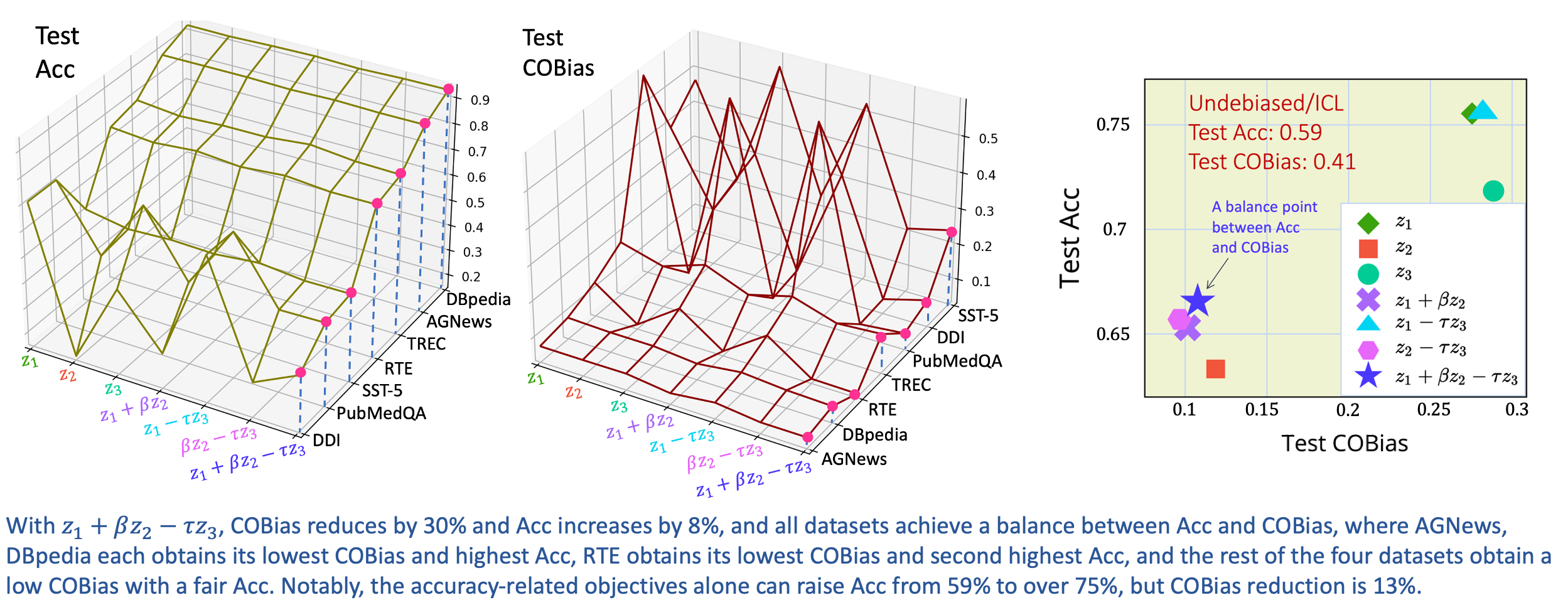}
 \caption{Ablation analysis; average scores over all datasets are used for the rightmost plot.}
\label{fig:ablation}
\end{figure*}

We show that \textbf{the integration of accuracy-related objectives and the COBias objective are indispensable} in achieving the best of both worlds in Figure \ref{fig:ablation}. DNIP objective function $z$ consists of three terms: the error rate term $z_1=\frac{1}{M} \sum_{m=1}^{M} \mathds{1}\{\hat{y}_m \ne y_m\}$; the COBias term $z_2={N\choose 2}^{-1} \sum_{i=1}^{N-1} \sum_{j=i+1}^{N} | A_{c_i} - A_{c_j} |$; and the PMI term $z_3= \sum_{j=1}^{N} \textrm{PMI}_j$; the accuracy-related terms are $z_1$ and $z_3$. We ablate objective functions to perform DNIP: $z_1, z_2, z_3, z_1+\beta z_2, z_1-\tau z_3, \beta z_2-\tau z_3, z_1+\beta z_2-\tau z_3$. For fair comparisons, we use 1-shot ICL with $\beta=2.7, \tau=0.2$, $K=30$. 

Among all ablations, $z_1+\beta z_2-\tau z_3$ achieves a balance point between accuracy and COBias. For the criteria, lower COBias is our top priority; when COBias scores are similar, we prefer the one with higher accuracy. Compared to $z_1+\beta z_2-\tau z_3$: (1) Using $z_1$ or $z_3$ solely, or $z_1-\tau z_3$, emphasizes improving accuracy only; (2) Using $z_2$ solely improves COBias but not accuracy; (3) Using $z_1+\beta z_2$ or $\beta z_2-\tau z_3$ is promising with a lower accuracy. Accuracy-related objectives alone increase accuracy by over 16\%, suggesting that DNIP is effective for accuracy improvement.

\section{Discussion}
\label{sec:discuss}
\noindent \textbf{Effect of Position Bias on DNIP.} 
Label ordering in few-shot prompts can introduce position bias, where outputs may favor patterns such as the majority label or the most recent label in the prompt \citep{zhao2021}, skewing class-wise accuracy. In our experiments, we applied three different seeds 0, 1, 2 to randomly select demonstrations used in the prompt.
\vspace{5pt}
\begin{wraptable}{r}{8.5cm}
\centering
\footnotesize
\setlength{\tabcolsep}{1.5pt}
\begin{tabular}{lccccccc}
\toprule
AGNews & \begin{tabular}[c]{@{}c@{}}Label\\ Order\end{tabular} & Acc. & COBias & \begin{tabular}[c]{@{}c@{}}World\\ (0)\end{tabular} & \begin{tabular}[c]{@{}c@{}}Sports\\ (1)\end{tabular} & \begin{tabular}[c]{@{}c@{}}Busi.\\ (2)\end{tabular} & \begin{tabular}[c]{@{}c@{}}Tech.\\ (3)\end{tabular} \\
\midrule
\multicolumn{8}{l}{5-shot ICL} \\
\midrule
seed0 & 3, 0, 0, 3, 3 & 84.6 & 20.5 & 89.4 & 98.7 & 91.5 & 58.5 \\
seed1 & 2, 1, 3, 2, 3 & 82.1 & 23.8 & 91.8 & 98.6 & 83.7 & 53.6 \\
seed2 & 0, 3, 3, 2, 3 & 80.6 & 28.9 & 89.0 & 98.8 & 92.3 & 42.0 \\
\midrule
\multicolumn{8}{l}{+DNIP} \\
\midrule
seed0 & - & 88.9 & 6.3 & 85.0 & 96.6 & 88.5 & 85.6 \\
seed1 & - & 88.8 & 6.7 & 86.5 & 97.4 & 86.8 & 84.2 \\
seed2 & - & 87.8 & 8.1 & 83.4 & 98.7 & 85.5 & 83.3 \\
\bottomrule
\end{tabular}
\caption{Label ordering of prompt demonstrations and test scores (\%).}
\label{tab:labelorder}
\end{wraptable}
For example, Table \ref{tab:labelorder} shows label ordering across three seeds for AGNews under 5-shot ICL. The ICL scores (Llama-2-13B) reflect variances caused by prompts over the three runs.

However, \textbf{the majority label (class 3 or 2) or the most recent label in the prompt (class 3) does not always receive the highest ICL accuracy}, rather, class 1 is the strongest across all seeds, suggesting that position bias does not play a role in our experimental results. \textbf{DNIP does not favor certain orders either.}\\

\noindent \textbf{Can Chain-of-Thought Prompting Reduce COBias?} Instead of post-hoc correction, Chain-of-Thought (CoT) \citep{cot} may elicit LLMs' reasoning capabilities for better generations. We additionally experiment with 1-shot CoT ICL on Gemma-2-2B, adding an example reasoning sentence for the demonstration and ``Answer: Think step by step before answering.'' to the prompt, while keeping other hyperparameters and prompt templates the same as the 1-shot ICL case. We extract the prediction (and class probabilities) by locating the last matching class token in the output; if none is found, we fall back to the last non-padding token.
\begin{wraptable}{r}{7.5cm}
\centering
\small
\resizebox{0.45\textwidth}{!}{
\begin{tabular}{lcccc}
\toprule
\begin{tabular}[c]{@{}l@{}}1-shot ICL\\ + CoT\end{tabular} & Accuracy & $\Delta$ & COBias & $\Delta$ \\ \midrule
AGN & 37.1\textsubscript{3.5} & {\color[HTML]{CB0000} -42.8} & 18.9\textsubscript{5.4} & {\color[HTML]{32CB00} -6.7} \\
DBP & 32.0\textsubscript{3.6} & {\color[HTML]{CB0000} -40.6} & 22.4\textsubscript{2.3} & {\color[HTML]{32CB00} -14.3} \\
SST & 23.2\textsubscript{1.0} & {\color[HTML]{CB0000} -4.4} & 29.1\textsubscript{1.1} & {\color[HTML]{32CB00} -12.9} \\
TREC & 20.3\textsubscript{2.4} & {\color[HTML]{CB0000} -24.3} & 25.5\textsubscript{2.2} & {\color[HTML]{32CB00} -24.7} \\
RTE & 48.5\textsubscript{1.6} & {\color[HTML]{CB0000} -21.8} & 36.4\textsubscript{7.1} & {\color[HTML]{CB0000} 11.2} \\
DDI & 35.4\textsubscript{7.9} & {\color[HTML]{32CB00} 19.9} & 29.7\textsubscript{1.4} & {\color[HTML]{32CB00} -5.4} \\
PMQA & 48.8\textsubscript{1.7} & {\color[HTML]{CB0000} -14.6} & 50.1\textsubscript{5.1} & {\color[HTML]{32CB00} -4.4} \\ \midrule
All & 35.0 & {\color[HTML]{CB0000} -18.4} & 30.3 & {\color[HTML]{32CB00} -8.2} \\ \bottomrule
\end{tabular}
}
\caption{1-shot CoT prompting. $\Delta$: absolute changes over 1-shot results. The drop in COBias is not due to better reasoning but due to noisier class probabilities.}
\label{tab:cot}
\end{wraptable}
Table \ref{tab:cot} shows that 1-shot CoT failed to maintain accuracy over vanilla 1-shot results, due to hallucinated continuations - matching label tokens are often missing, so class probabilities at the expected answer position are noisier or more randomly distributed. This leads to \textbf{more uniform but unreliable class predictions, lowering COBias at the cost of accuracy}. While careful prompt engineering for CoT (more reasoning steps, etc.) may reduce hallucinations, these strategies require significant GPU resources at inference. In contrast, DNIP operates post-hoc on LLM outputs with CPU resources, effectively debiasing predictions from 1-shot ICL without modifying prompts or GPU consumptions.

\section{Conclusion, Limitations, \& Future Work}
The COBias metric and the class-level debiasing method DNIP address the problem of imbalanced class accuracy in LLMs outputs. Our method not only offers a robust post-hoc debiasing solution, but also shows a new path to directly build hard-to-optimize accuracy goals into how LLMs learn, powered by integer optimization.

While DNIP's probability correction through reweighting coefficients is straightforward and easy to plug in, its main limit is that it works only at the class level. This means it applies the same correction to all samples within an output class, so it can not tailor the probability correction to the unique characteristics of different individual samples. For future work, a more refined debiasing could utilize correction functions to reweight class probabilities at a finer granularity. The sample-level debiasing will account for the heterogeneous probability distributions of individual samples, leading to more precise and tailored adjustments, especially elevating weaker classes.

\bibliography{output}{}
\bibliographystyle{unsrtnat}

\newpage
\appendix

\section{A. The Odd Class Phenomenon and $\text{COBias}_{\text{single}}$}
\label{appdix:a}
\label{sec:cobiassingle}
While our primary metric for overall class accuracy imbalance is $\text{COBias}$, we also utilize $\text{COBias}_{\text{single}}$ to specifically capture biases arising from dominant misprediction patterns, a phenomenon we term the `odd class'. This section provides a detailed definition of this concept and the $\text{COBias}_{\text{single}}$ metric.

\begin{table}[ht]
    \centering
    \small
    \begin{center}
  \begin{tabular}{R{2cm} R{0.6cm} R{0.6cm}R{0.8cm} R{0.5cm}R{0.4cm}R{0.4cm}}
  \toprule
    \textbf{} &  &  & \textbf{\textit{Pred.}} &  & & \\
    & \textbf{World} & \textbf{Sports} & \textbf{Business} & \textbf{Tech} & \textbf{All} & \textbf{Acc.} \\
    \midrule
    \textbf{World} & 1093 & 64 & 126 & 3 & 1286 & 0.85\\
    \textbf{Sports} & 9 & 1247 & 14 & 0 & 1270& 0.98\\
    \textbf{\textit{True}} \textbf{Business} & 25 & 4 & 1167 & 8 & \textbf{\color{red}1204\color{black}}& \textbf{\color{red}0.97\color{black}} \\
    \textbf{Tech} & 156 & 27 & \textbf{\color{red}822\color{black}} & 235 & \textbf{\color{red}1240\color{black}} & \textbf{\color{red}0.19\color{black}}\\
    \textbf{All} & 1283 & 1342 & \textbf{\color{red}2129\color{black}} & \textbf{\color{red}246\color{black}} & 5000 & -\\
    \textbf{Prec.} & 0.85 & 0.93 & 0.55 & 0.96 & - & -\\
    \bottomrule
    \end{tabular}
    \end{center}
\caption{Over-prediction in class \textit{Business} and under-prediction in class \textit{Tech} for AGNews in Llama-2-13B, shown by the test confusion matrix.}
\label{tab:llama213b_ag}
\end{table}

\subsection{A.1 Conceptualizing the Odd Class}
In instances of class accuracy imbalance, we often observe that a ground-truth class is disproportionately misclassified as a specific \textit{other} class. This specific output class, which receives the most wrong predictions for a given ground-truth class, is what we define as the `odd class'. For $N$ classes and $M$ instances, let $y_m$ denote the ground-truth class and $\hat{y}_m$ the prediction for instance $m$. The number of ground-truth $c_i$ instances wrongly predicted as $c_j$ is given by:
\begin{equation}
    u_{c_i,c_j} = \sum_{m=1}^{M} \mathbf{1}\{y_m = i, \hat{y}_m = j\} \quad \text{for } j \neq i
\end{equation}
The odd class for a ground-truth class $c_i$, denoted $\text{odd}_i$, is then formally defined as:
\begin{equation}
    \text{odd}_i = \text{odd class}(c_i) := \arg \max_{c_j} u_{c_i,c_j}
\end{equation}
For example, as highlighted in Table \ref{tab:llama213b_ag} (copied from the main paper), for ground-truth \textit{Tech}, the most significant contributor to its under-prediction is the 822 instances wrongly predicted as \textit{Business}, making \textit{Business} the $\text{odd class for \textit{Tech}}$.

\begin{wrapfigure}[12]{R}{0.45\textwidth}
    \centering
\includegraphics[width=0.7\linewidth]{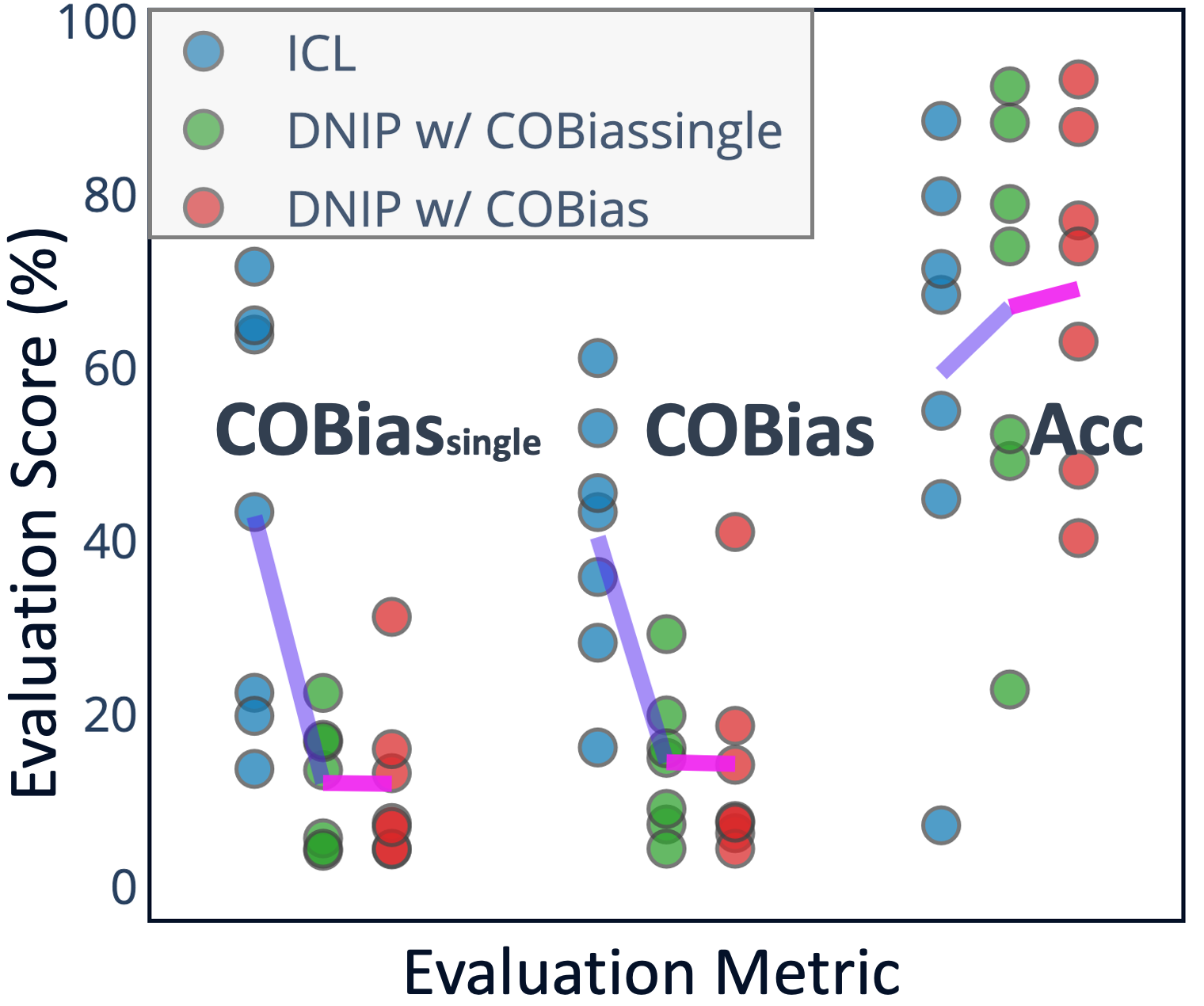}
    \caption{Test scores when optimizing COBias\textsubscript{single}.}
  \label{fig:cobiassingle}
\end{wrapfigure}

\subsection{A.2 Definition of $\text{COBias}_{\text{single}}$}
Building upon the odd class concept, $\text{COBias}_{\text{single}}$ quantifies how much an odd class contributes to the under-prediction of its corresponding ground-truth class. It is computed as the average absolute accuracy difference between each ground-truth class and its identified odd class:
\begin{equation}
\text{COBias}_{\text{single}} = \frac{1}{N} \sum_{i=1}^{N} |A_{\text{odd}_i} - A_{c_i}| 
\label{eq:cbsingle}
\end{equation}
where $A_{c_i}$ is the accuracy for class $c_i$, and $A_{\text{odd}_i}$ is the accuracy for its odd class $\text{odd}_i$. $\text{COBias}_{\text{single}}$ measures a model’s tendency to under-predict the true answer and mostly mistake it as an odd answer. The term "contextual" in Contextual Oddity Bias highlights that the identity of the odd class can vary depending on the specific model or task.

\subsection{A.3 Evaluation of $\text{COBias}_{\text{single}}$}
Using COBias\textsubscript{single} for DNIP optimization will focus on odd classes only. Figure \ref{fig:cobiassingle} shows test scores of the three evaluation metrics when optimizing DNIP with the COBias term replaced by COBias\textsubscript{single} in the objective function. Blue dots represent 1-shot ICL scores (Llama-2-13B) of each dataset; green dots are DNIP results using the COBias\textsubscript{single} objective; red dots use the original objective. 

DNIP with COBias\textsubscript{single} significantly minimizes accuracy differences between odd classes and ground-truth classes, but DNIP with COBias is even more effective as it balances all pairs of classes.

\section{B. An Illustration of the Process of DNIP}
\label{appdix:b}
\begin{figure*}[htb]
  \centering
  \includegraphics[width=\linewidth]{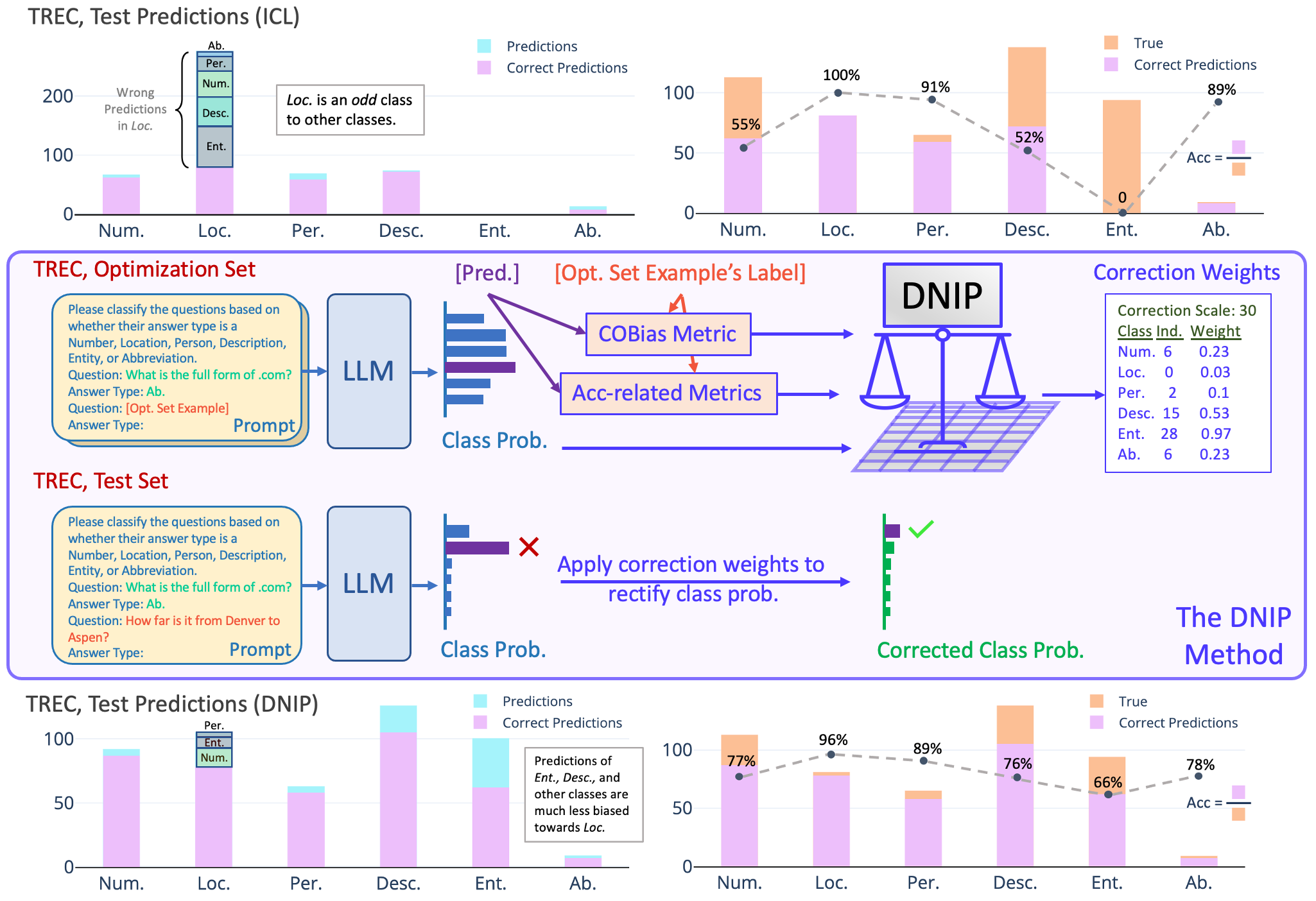}
  \caption{An overview of the DNIP method using the TREC task for illustration; the ICL approach uses 1-shot prompting whose prompt is given by light yellow text boxes. The upper-left plot shows that \textit{Loc.} is an odd class to the others, holding their majority wrong predictions, especially for \textit{Ent.}; the upper-right plot demonstrates the stark contrast in class accuracy, with 100\% for \textit{Loc.} and 0 for \textit{Ent.}. The bottom plots show that DNIP greatly reduces pairwise class accuracy differences, especially boosting accuracy for \textit{Ent.} ($0 \rightarrow 66\%$).}
  \label{fig:overview}
\end{figure*}

Figure \ref{fig:overview} illustrate the DNIP optimization and correction process with the TREC dataset.

\section{C. Derivation of SA Algorithm Complexity and Proof for Convergence}
\label{appdix:c}
Given $N$ classes and $K$ scales, the variable space size is $NK$. For each inner loop of SA, the length of the Markov chain is often several times of the variable space, i.e., $\lambda NK$. For a geometric cooling schedule $T = \alpha^{n} T\textsubscript{max}$, the number of outer-loop iterations is:
\begin{equation}
    N^{\text{outer}} = \log_{\alpha}{\frac{T\textsubscript{min}}{T\textsubscript{max}}}
\end{equation}

The sampling times for the SA algorithm is $\lambda NK \log_{\alpha}{(T\textsubscript{min}/T\textsubscript{max})}$. Therefore, the computational complexity is $\mathcal{O}(\lambda NK \log_{\alpha}{(T\textsubscript{min}/T\textsubscript{max})})=\mathcal{O}(NK)$.

In addition, the brute force strategy to find optimal correction weights is enumerated search, where the number of weight combinations to explore is $K^{N}$ for an example; for $M$ examples, it results in a computational complexity of $\mathcal{O}(K^{N} \cdot M)$, which can not finish in polynomial time. When $M$ or $N$ is large, enumerated search is computationally intensive. Considering this, we did not adopt the brute force strategy.

\noindent \textbf{Convergence of SA with Search Complexity $\mathcal{O}(NK)$ and Geometric Cooling}\\
We show how the simulated annealing (SA) algorithm used in DNIP is guaranteed to probabilistically converge to global optima. To recall, in our DNIP optimization, class $i$ selects a weight index $\xi_i$ from a $K$-point scale $\boldsymbol{\omega}$ to correct its probability. To simply notations, let $w_i=\boldsymbol{\omega}_{\xi_i}$ be the correction weight for class $i$. Then, we denote $S=\{w \in \mathbbm{R}^{N}|w_i \in \{1/K,2/K,\dots,1\}\}$ as the search space of correction weights $w=(w_1,\dots,w_N)$. We also rewrite the objective function as $Z(w)$. Therefore, the optimal weights satisfy 
\begin{equation}
    w^{\ast}=\argmin_{w \in S} Z(w).
\end{equation}
In addition, let $S^{\ast}$ be the set of globally optimal solutions.

SA explores $S$ using a Markov chain at time $t$, with transition probability $P(w,w',t)$ from $w$ to $w'$ following the Metropolis-Hastings acceptance probability:
\begin{equation}
P(w,w',t)
=\begin{cases}
q(w,w')a(w,w',t) & w \ne w'\\
1-\sum_{\substack{w'' \in S, w'' \ne w}} P(w,w'',t) & w=w'
\end{cases}   
\end{equation}
where $q(w,w')$ is a proposal distribution that randomly selects a class $i$ and perturbs $w_i$, scaling as $\mathcal{O}(NK)$. Under this search complexity, the Markov chain will eventually visit all states with sufficiently long execution; $a(w,w',t)=\exp{(\frac{Z(w)-Z(w')}{T(t)})}$ is the acceptance probability with temperature schedule $T(t)$. To simplify the notation, the transition probability is given by:
\begin{equation}
    P(w,w',t)=q(w,w')A(w,w',t)
\end{equation}
where $A(w,w',t) = \min  (1,\exp{(\frac{Z(w)-Z(w')}{T(t)})})$.

The Metropolis-Hastings algorithm generates a Markov chain that satisfies the detailed balance condition \citep{Hastings1970}. That is, the probability of being in state $w$ at time $t$ satisfies:
\begin{equation}
    P(w,t)q(w,w')A(w,w',t)=P(w',t)q(w',w)A(w',w,t).
\end{equation}
At thermodynamic equilibrium, the Markov chain follows a Boltzmann distribution \citep{kirkpatrick1983}, where
\begin{equation}
    P(w,t) \propto \exp{\left(-\frac{Z(w)}{T(t)}\right)}.
\end{equation}
Then, the the probability ratio between a suboptimal state (solution) $w$ and the optimal state $w^{\ast}$ is:
\begin{equation}
    \frac{P(w,t)}{P(w^{\ast},t)}=\exp{\left(\frac{Z(w^{\ast})-Z(w)}{T(t)}\right)}.
\end{equation}
Substituting our geometric cooling schedule \citep{hajek1988} with a cooling rate $c$ and initial temperature $T_0$ ($c$ is close to 1 to make the cooling slow enough):
\begin{equation}
 T(t+1)=c \cdot T(t)=T_{0} c^t, \qquad 0<c<1,  
\end{equation}
the above ratio becomes
\begin{equation}
    \frac{P(w,t)}{P(w^{\ast},t)}=\exp{\left(\frac{Z(w^{\ast})-Z(w)}{T_{0} c^t}\right)}.
\end{equation}
Taking the limit as $t \rightarrow \infty$, since $T_{0} c^t \rightarrow 0, Z(w^{\ast})-Z(w)<0$, we obtain
\begin{equation}
    \lim\limits_{t \to \infty} \frac{P(w,t)}{P(w^{\ast},t)}=\lim\limits_{t \to \infty} \exp{\left(\frac{Z(w^{\ast})-Z(w)}{T_{0} c^t}\right)}=0.
\end{equation}
Therefore,
\begin{equation}
    \lim\limits_{t \to \infty} P(w,t)=0, \qquad \forall w \notin S^{\ast}.
\end{equation}
Now, we sum over all suboptimal states and take the limit, we can obtain
\begin{equation}
    \lim\limits_{t \to \infty} P_{\text{so}}(t)=\lim\limits_{t \to \infty}\sum\limits_{w \notin S^{\ast}} P(w,t)=\sum\limits_{w \notin S^{\ast}} \lim\limits_{t \to \infty}P(w,t)=0.
\end{equation}
This shows that as $t \rightarrow \infty$, the probability of being in a suboptimal solution goes to 0. It proves that the Markov chain converges to an optimal solution.

In our experiments, the annealing process uses a geometric cooling schedule with a decay $\alpha = 0.95$ and an initial temperature of 200,000. The algorithm's inner loop simulates thermal equilibrium, with a stopping criterion based on the number of generated or accepted solutions \citep{fabio1991}.

\section{D. DNIP Using Smaller Optimization Sets}
\label{appdix:d}
We evaluate DNIP with varying optimization examples (GPT-2-XL, Llama-2-7B/13B). DNIP requires only a small optimization set of 10 examples for a fair improvement in both COBias and accuracy, and it exhibits a further, emergent COBias reduction at several thousands optimization examples. The full results for three models across varying optimization set sizes are shown in Table \ref{tab:varytrain_full}.

\begin{table*}[htbp!]
\setlength\tabcolsep{3.6pt}
\centering
\begin{tabular}{ccccccccc}
\toprule
\multicolumn{1}{c}{Opt. set} & \multicolumn{1}{c}{Eval. Metric} & \multicolumn{1}{c}{AGNews}  & \multicolumn{1}{c}{DBpedia}  & \multicolumn{1}{c}{SST-5} & \multicolumn{1}{c}{TREC} & \multicolumn{1}{c}{RTE} & \multicolumn{1}{c}{DDI} & \multicolumn{1}{c}{PubMedQA}  \\
\midrule
\multicolumn{9}{c}{\textit{GPT-2-XL}}\\
\midrule
0 &
\begin{tabular}[c]{@{}c@{}}Acc\\ \color{blue}COBias\end{tabular}     &
\begin{tabular}[c]{@{}l@{}}52.1\textsubscript{5.4}\\ \color{blue}35.5\textsubscript{11.5}\end{tabular}     &
\begin{tabular}[c]{@{}l@{}}31.8\textsubscript{9.9}\\ \color{blue}40.0\textsubscript{3.6}\end{tabular} & 
\begin{tabular}[c]{@{}l@{}}34.9\textsubscript{13.7}\\ \color{blue}48.7\textsubscript{5.4}\end{tabular} &
\begin{tabular}[c]{@{}l@{}}27.4\textsubscript{10.5}\\ \color{blue}45.6\textsubscript{8.7}\end{tabular} &
\begin{tabular}[c]{@{}l@{}}\textbf{55.4}\textsubscript{1.9}\\ \color{blue}82.4\textsubscript{24.5}\end{tabular} & 
\begin{tabular}[c]{@{}l@{}}14.5\textsubscript{4.4}\\ \color{blue}40.7\textsubscript{5.9}\end{tabular} &  
\begin{tabular}[c]{@{}l@{}}55.2\textsubscript{0.0}\\ \color{blue}59.4\textsubscript{12.6}\end{tabular}  \\ 
10\textsuperscript{$\ast$}& 
\begin{tabular}[c]{@{}c@{}}Acc\\ \color{blue}COBias\end{tabular} &
\begin{tabular}[c]{@{}l@{}}\textbf{72.8}\textsubscript{2.1}\\ \color{blue}17.2\textsubscript{1.7}\end{tabular} & 
\begin{tabular}[c]{@{}l@{}}40.6\textsubscript{5.2}\\ \color{blue}28.0\textsubscript{6.6}\end{tabular} & 
\begin{tabular}[c]{@{}l@{}}40.3\textsubscript{3.4}\\ \color{blue}40.1\textsubscript{1.8}\end{tabular} & 
\begin{tabular}[c]{@{}l@{}}\textbf{51.5}\textsubscript{5.7}\\ \color{blue}39.7\textsubscript{1.0}\end{tabular} & 
\begin{tabular}[c]{@{}l@{}}50.3\textsubscript{0.8}\\ \color{blue}6.5\textsubscript{7.6}\end{tabular} & 
\begin{tabular}[c]{@{}l@{}}23.5\textsubscript{1.0}\\ \color{blue}40.6\textsubscript{3.5}\end{tabular} & 
\begin{tabular}[c]{@{}l@{}}53.7\textsubscript{0.6}\\ \color{blue}41.8\textsubscript{0.0}\end{tabular}     \\ 
100 & 
\begin{tabular}[c]{@{}c@{}}Acc\\ \color{blue}COBias\end{tabular} &
\begin{tabular}[c]{@{}l@{}}70.0\textsubscript{5.0}\\ \color{blue}14.8\textsubscript{1.3}\end{tabular} & 
\begin{tabular}[c]{@{}l@{}}41.4\textsubscript{29.9}\\ \color{blue}27.8\textsubscript{8.9}\end{tabular} & 
\begin{tabular}[c]{@{}l@{}}44.3\textsubscript{1.3}\\ \color{blue}30.8\textsubscript{3.4}\end{tabular} & 
\begin{tabular}[c]{@{}l@{}}49.0\textsubscript{10.1}\\ \color{blue}31.1\textsubscript{5.9}\end{tabular} & 
\begin{tabular}[c]{@{}l@{}}49.8\textsubscript{2.5}\\ \color{blue}5.3\textsubscript{4.9}\end{tabular} & 
\begin{tabular}[c]{@{}l@{}}42.2\textsubscript{5.7}\\ \color{blue}17.1\textsubscript{7.1}\end{tabular} & 
\begin{tabular}[c]{@{}l@{}}46.7\textsubscript{10.8}\\ \color{blue}29.2\textsubscript{16.6}\end{tabular}  \\ 
500 & 
\begin{tabular}[c]{@{}c@{}}Acc\\ \color{blue}COBias\end{tabular} &
\begin{tabular}[c]{@{}l@{}}68.2\textsubscript{1.2}\\ \color{blue}2.7\textsubscript{1.8}\end{tabular} & 
\begin{tabular}[c]{@{}l@{}}44.5\textsubscript{33.2}\\ \color{blue}25.7\textsubscript{9.6}\end{tabular} & 
\begin{tabular}[c]{@{}l@{}}43.4\textsubscript{1.5}\\ \color{blue}29.8\textsubscript{3.7}\end{tabular} & 
\begin{tabular}[c]{@{}l@{}}49.9\textsubscript{8.5}\\ \color{blue}31.4\textsubscript{6.2}\end{tabular} & 
\begin{tabular}[c]{@{}l@{}}50.5\textsubscript{2.4}\\ \color{blue}\textbf{5.2}\textsubscript{3.9}\end{tabular} & 
\begin{tabular}[c]{@{}l@{}}45.6\textsubscript{9.2}\\ \color{blue}24.5\textsubscript{14.3}\end{tabular} & 
\begin{tabular}[c]{@{}l@{}}49.9\textsubscript{11.4}\\ \color{blue}\textbf{17.7}\textsubscript{15.8}\end{tabular}  \\ 
1,000 & 
\begin{tabular}[c]{@{}c@{}}Acc\\ \color{blue}COBias\end{tabular} &
\begin{tabular}[c]{@{}l@{}}69.6\textsubscript{1.0}\\ \color{blue}3.0\textsubscript{1.2}\end{tabular} & 
\begin{tabular}[c]{@{}l@{}}42.0\textsubscript{32.2}\\ \color{blue}27.6\textsubscript{11.0}\end{tabular} & 
\begin{tabular}[c]{@{}l@{}}42.8\textsubscript{2.3}\\ \color{blue}28.2\textsubscript{5.9}\end{tabular} & 
\begin{tabular}[c]{@{}l@{}}50.9\textsubscript{9.0}\\ \color{blue}\textbf{24.5}\textsubscript{5.4}\end{tabular} & 
\begin{tabular}[c]{@{}l@{}}50.3\textsubscript{2.2}\\ \color{blue}\textbf{5.2}\textsubscript{3.3}\end{tabular} & 
\begin{tabular}[c]{@{}l@{}}48.3\textsubscript{4.9}\\ \color{blue}22.9\textsubscript{2.5}\end{tabular} & 
\begin{tabular}[c]{@{}c@{}}same as\\ full\end{tabular}  \\  
Full & 
\begin{tabular}[c]{@{}c@{}}Acc\\ \color{blue}COBias\end{tabular}     &
\begin{tabular}[c]{@{}l@{}}68.5\textsubscript{1.0}\\ \color{blue}\textbf{1.4}\textsubscript{0.5}\end{tabular}     &
\begin{tabular}[c]{@{}l@{}}\textbf{69.9}\textsubscript{9.1}\\ \color{blue}\textbf{24.1}\textsubscript{8.3}\end{tabular} & 
\begin{tabular}[c]{@{}l@{}}\textbf{44.5}\textsubscript{2.20}\\ \color{blue}\textbf{26.0}\textsubscript{2.5}\end{tabular} &
\begin{tabular}[c]{@{}l@{}}46.3\textsubscript{12.7}\\ \color{blue}27.2\textsubscript{7.2}\end{tabular} &
\begin{tabular}[c]{@{}l@{}}50.8\textsubscript{2.1}\\ \color{blue}7.1\textsubscript{5.0}\end{tabular} & 
\begin{tabular}[c]{@{}l@{}}43.9\textsubscript{14.9}\\ \color{blue}\textbf{17.0}\textsubscript{7.1}\end{tabular} &  
\begin{tabular}[c]{@{}l@{}}\textbf{57.1}\textsubscript{1.3}\\ \color{blue}29.8\textsubscript{25.0}\end{tabular} \\ 
\midrule
\multicolumn{9}{c}{\textit{Llama-2-7B}}\\
\midrule
0 &
\begin{tabular}[c]{@{}c@{}}Acc\\ \color{blue}COBias\end{tabular} &
\begin{tabular}[c]{@{}l@{}}86.4\textsubscript{2.5}\\ \color{blue}14.0\textsubscript{6.5}\end{tabular}     &
\begin{tabular}[c]{@{}l@{}}88.9\textsubscript{2.0}\\ \color{blue}13.5\textsubscript{2.1}\end{tabular} & 
\begin{tabular}[c]{@{}l@{}}42.1\textsubscript{11.1}\\ \color{blue}55.6\textsubscript{1.5}\end{tabular} &
\begin{tabular}[c]{@{}l@{}}66.7\textsubscript{6.6}\\ \color{blue}33.2\textsubscript{10.0}\end{tabular} &
\begin{tabular}[c]{@{}l@{}}66.3\textsubscript{4.3}\\ \color{blue}61.6\textsubscript{10.5}\end{tabular} & 
\begin{tabular}[c]{@{}l@{}}6.7\textsubscript{0.4}\\ \color{blue}41.4\textsubscript{1.7}\end{tabular} &  
\begin{tabular}[c]{@{}l@{}}40.3\textsubscript{6.7}\\ \color{blue}40.9\textsubscript{16.1}\end{tabular}     \\
10\textsuperscript{$\ast$}& 
\begin{tabular}[c]{@{}c@{}}Acc\\ \color{blue}COBias\end{tabular} &
\begin{tabular}[c]{@{}l@{}}86.4\textsubscript{2.5}\\ \color{blue}14.0\textsubscript{6.5}\end{tabular} & 
\begin{tabular}[c]{@{}l@{}}89.9\textsubscript{1.4}\\ \color{blue}12.5\textsubscript{2.1}\end{tabular} & 
\begin{tabular}[c]{@{}l@{}}\textbf{51.4}\textsubscript{0.9}\\ \color{blue}36.2\textsubscript{3.6}\end{tabular} & 
\begin{tabular}[c]{@{}l@{}}\textbf{70.1}\textsubscript{0.9}\\ \color{blue}22.4\textsubscript{5.4}\end{tabular} & 
\begin{tabular}[c]{@{}l@{}}\textbf{74.9}\textsubscript{2.1}\\ \color{blue}4.8\textsubscript{5.1}\end{tabular} & 
\begin{tabular}[c]{@{}l@{}}31.7\textsubscript{21.0}\\ \color{blue}26.7\textsubscript{8.2}\end{tabular} &
\begin{tabular}[c]{@{}l@{}}44.5\textsubscript{0.6}\\ \color{blue}\textbf{28.7}\textsubscript{3.9}\end{tabular} \\ 
100 & 
\begin{tabular}[c]{@{}c@{}}Acc\\ \color{blue}COBias\end{tabular} &
\begin{tabular}[c]{@{}l@{}}\textbf{88.4}\textsubscript{0.4}\\ \color{blue}5.8\textsubscript{0.6}\end{tabular} & 
\begin{tabular}[c]{@{}l@{}}91.8\textsubscript{0.7}\\ \color{blue}9.7\textsubscript{1.0}\end{tabular} & 
\begin{tabular}[c]{@{}l@{}}50.9\textsubscript{1.5}\\ \color{blue}34.3\textsubscript{12.7}\end{tabular} & 
\begin{tabular}[c]{@{}l@{}}\textbf{70.1}\textsubscript{1.0}\\ \color{blue}16.7\textsubscript{2.4}\end{tabular} & 
\begin{tabular}[c]{@{}l@{}}73.6\textsubscript{2.6}\\ \color{blue}2.9\textsubscript{0.7}\end{tabular} & 
\begin{tabular}[c]{@{}l@{}}44.9\textsubscript{2.5}\\ \color{blue}21.0\textsubscript{6.6}\end{tabular} &
\begin{tabular}[c]{@{}l@{}}62.6\textsubscript{4.5}\\ \color{blue}35.4\textsubscript{18.1}\end{tabular}  \\ 
500 & 
\begin{tabular}[c]{@{}c@{}}Acc\\ \color{blue}COBias\end{tabular} &
\begin{tabular}[c]{@{}l@{}}86.8\textsubscript{0.9}\\ \color{blue}2.8\textsubscript{0.3}\end{tabular} & 
\begin{tabular}[c]{@{}l@{}}92.1\textsubscript{0.6}\\ \color{blue}8.6\textsubscript{1.6}\end{tabular} & 
\begin{tabular}[c]{@{}l@{}}50.8\textsubscript{1.7}\\ \color{blue}35.8\textsubscript{12.3}\end{tabular} & 
\begin{tabular}[c]{@{}l@{}}69.7\textsubscript{1.2}\\ \color{blue}\textbf{15.2}\textsubscript{2.1}\end{tabular} & 
\begin{tabular}[c]{@{}l@{}}74.3\textsubscript{2.9}\\ \color{blue}3.1\textsubscript{1.4}\end{tabular} & 
\begin{tabular}[c]{@{}l@{}}\textbf{69.3}\textsubscript{1.7}\\ \color{blue}34.5\textsubscript{0.3}\end{tabular} &
\begin{tabular}[c]{@{}l@{}}\textbf{63.5}\textsubscript{7.5}\\ \color{blue}37.3\textsubscript{20.4}\end{tabular} \\ 
1,000 & 
\begin{tabular}[c]{@{}c@{}}Acc\\ \color{blue}COBias\end{tabular} &
\begin{tabular}[c]{@{}l@{}}86.8\textsubscript{0.3}\\ \color{blue}1.9\textsubscript{0.5}\end{tabular} & 
\begin{tabular}[c]{@{}l@{}}92.5\textsubscript{0.2}\\ \color{blue}8.0\textsubscript{0.4}\end{tabular} & 
\begin{tabular}[c]{@{}l@{}}51.0\textsubscript{2.1}\\ \color{blue}30.9\textsubscript{20.4}\end{tabular} & 
\begin{tabular}[c]{@{}l@{}}69.5\textsubscript{0.8}\\ \color{blue}15.7\textsubscript{1.2}\end{tabular} & 
\begin{tabular}[c]{@{}l@{}}74.0\textsubscript{2.5}\\ \color{blue}2.6\textsubscript{1.2}\end{tabular} & 
\begin{tabular}[c]{@{}l@{}}56.5\textsubscript{7.9}\\ \color{blue}29.5\textsubscript{3.4}\end{tabular} & 
\begin{tabular}[c]{@{}c@{}}same as\\ full\end{tabular}  \\  
Full & 
\begin{tabular}[c]{@{}c@{}}Acc\\ \color{blue}COBias\end{tabular}     &
\begin{tabular}[c]{@{}l@{}}86.7\textsubscript{0.4}\\ \color{blue}\textbf{1.3}\textsubscript{0.1}\end{tabular}     &
\begin{tabular}[c]{@{}l@{}}\textbf{92.9}\textsubscript{0.4}\\ \color{blue}\textbf{7.7}\textsubscript{0.6}\end{tabular} & 
\begin{tabular}[c]{@{}l@{}}50.6\textsubscript{2.7}\\ \color{blue}\textbf{28.0}\textsubscript{21.6}\end{tabular} &
\begin{tabular}[c]{@{}l@{}}68.1\textsubscript{1.0}\\ \color{blue}15.9\textsubscript{1.6}\end{tabular} &
\begin{tabular}[c]{@{}l@{}}73.9\textsubscript{2.3}\\ \color{blue}\textbf{1.9}\textsubscript{1.8}\end{tabular} & 
\begin{tabular}[c]{@{}l@{}}44.5\textsubscript{3.8}\\ \color{blue}\textbf{11.6}\textsubscript{3.0}\end{tabular} &  
\begin{tabular}[c]{@{}l@{}}62.7\textsubscript{8.3}\\ \color{blue}35.4\textsubscript{22.8}\end{tabular}     \\ 
\midrule
\multicolumn{9}{c}{\textit{Llama-2-13B}}\\
\midrule
0 &
\begin{tabular}[c]{@{}c@{}}Acc\\ \color{blue}COBias\end{tabular}     &
\begin{tabular}[c]{@{}l@{}}79.9\textsubscript{7.0}\\ \color{blue}28.3\textsubscript{16.1}\end{tabular}     &
\begin{tabular}[c]{@{}l@{}}88.6\textsubscript{1.7}\\ \color{blue}16.2\textsubscript{3.7}\end{tabular} & 
\begin{tabular}[c]{@{}l@{}}44.9\textsubscript{4.3}\\ \color{blue}53.1\textsubscript{5.0}\end{tabular} &
\begin{tabular}[c]{@{}l@{}}68.5\textsubscript{10.8}\\ \color{blue}35.9\textsubscript{6.5}\end{tabular} &
\begin{tabular}[c]{@{}l@{}}71.5\textsubscript{2.2}\\ \color{blue}43.4\textsubscript{7.0}\end{tabular} & 
\begin{tabular}[c]{@{}l@{}}7.2\textsubscript{0.9}\\ \color{blue}45.6\textsubscript{5.9}\end{tabular} &  
\begin{tabular}[c]{@{}l@{}}55.1\textsubscript{2.9}\\ \color{blue}61.2\textsubscript{1.9}\end{tabular} \\
10\textsuperscript{$\ast$}& 
\begin{tabular}[c]{@{}c@{}}Acc\\ \color{blue}COBias\end{tabular} &
\begin{tabular}[c]{@{}l@{}}86.0\textsubscript{1.9}/\\\color{blue}14.3\textsubscript{3.5}\end{tabular} & 
\begin{tabular}[c]{@{}l@{}}88.8\textsubscript{1.6}/\\\color{blue}14.7\textsubscript{2.5}\end{tabular} & 
\begin{tabular}[c]{@{}l@{}}49.2\textsubscript{0.8}/\\\color{blue}41.7\textsubscript{8.1}\end{tabular} & 
\begin{tabular}[c]{@{}l@{}}75.8\textsubscript{3.2}\\ \color{blue}	29.6\textsubscript{4.5}\end{tabular} & 
\begin{tabular}[c]{@{}l@{}}74.8\textsubscript{2.3}\\ \color{blue}12.2\textsubscript{5.7}\end{tabular} & 
\begin{tabular}[c]{@{}l@{}}18.7\textsubscript{12.6}\\ \color{blue}30.0\textsubscript{7.8}\end{tabular} & 
\begin{tabular}[c]{@{}l@{}}59.6\textsubscript{6.2}\\ \color{blue}35.3\textsubscript{7.8}\end{tabular}   \\ 
100 & 
\begin{tabular}[c]{@{}c@{}}Acc\\ \color{blue}COBias\end{tabular} &
\begin{tabular}[c]{@{}l@{}}87.8\textsubscript{0.2}/\\\color{blue}8.3\textsubscript{0.5}\end{tabular} & 
\begin{tabular}[c]{@{}l@{}}91.8\textsubscript{0.2}/\\\color{blue}10.1\textsubscript{0.7}\end{tabular} & 
\begin{tabular}[c]{@{}l@{}}47.4\textsubscript{2.0}/\\\color{blue}30.4\textsubscript{4.4}\end{tabular} & 
\begin{tabular}[c]{@{}l@{}}78.3\textsubscript{2.8}\\ \color{blue}18.5\textsubscript{2.6}\end{tabular} & 
\begin{tabular}[c]{@{}l@{}}\textbf{75.5}\textsubscript{1.0}\\ \color{blue}10.4\textsubscript{6.9}\end{tabular} & 
\begin{tabular}[c]{@{}l@{}}22.2\textsubscript{2.0}\\ \color{blue}17.3\textsubscript{4.4}\end{tabular} & 
\begin{tabular}[c]{@{}l@{}}56.9\textsubscript{10.4}\\ \color{blue}25.8\textsubscript{16.3}\end{tabular}  \\ 
500 & 
\begin{tabular}[c]{@{}c@{}}Acc\\ \color{blue}COBias\end{tabular} &
\begin{tabular}[c]{@{}l@{}}87.1\textsubscript{1.5}\\ \color{blue}\textbf{5.3}\textsubscript{3.1}\end{tabular} & 
\begin{tabular}[c]{@{}l@{}}92.7\textsubscript{0.0}/\color{blue}\\9.1\textsubscript{0.3}\end{tabular} & 
\begin{tabular}[c]{@{}l@{}}49.5\textsubscript{0.7}/\color{blue}\\28.0\textsubscript{3.2}\end{tabular} & 
\begin{tabular}[c]{@{}l@{}}\textbf{78.7}\textsubscript{2.1}\\ \color{blue}16.3\textsubscript{1.0}\end{tabular} & 
\begin{tabular}[c]{@{}l@{}}73.6\textsubscript{1.7}\\ \color{blue}4.0\textsubscript{3.4}\end{tabular} & 
\begin{tabular}[c]{@{}l@{}}48.2\textsubscript{27.4}\\ \color{blue}28.0\textsubscript{11.5}\end{tabular} & 
\begin{tabular}[c]{@{}l@{}}62.2\textsubscript{15.8}\\ \color{blue}40.4\textsubscript{29.4}\end{tabular}   \\ 
1,000 & 
\begin{tabular}[c]{@{}c@{}}Acc\\ \color{blue}COBias\end{tabular} &
\begin{tabular}[c]{@{}l@{}}87.7\textsubscript{0.7}/\\\color{blue}6.7\textsubscript{1.5}\end{tabular} & 
\begin{tabular}[c]{@{}l@{}}92.8\textsubscript{0.1}/\\\color{blue}9.2\textsubscript{0.2}\end{tabular} & 
\begin{tabular}[c]{@{}l@{}}\textbf{49.8}\textsubscript{0.7}/\\\color{blue}28.8\textsubscript{2.0}\end{tabular} & 
\begin{tabular}[c]{@{}l@{}}77.3\textsubscript{2.0}\\ \color{blue}\textbf{14.2}\textsubscript{1.8}\end{tabular} & 
\begin{tabular}[c]{@{}l@{}}73.9\textsubscript{1.5}\\ \color{blue}\textbf{3.6}\textsubscript{4.3}\end{tabular} & 
\begin{tabular}[c]{@{}l@{}}\textbf{52.0}\textsubscript{10.5}\\ \color{blue}25.0\textsubscript{9.7}\end{tabular} & 
\begin{tabular}[c]{@{}c@{}}same as\\ full\end{tabular}   \\ 
Full & 
\begin{tabular}[c]{@{}c@{}}Acc\\ \color{blue}COBias\end{tabular}     &
\begin{tabular}[c]{@{}l@{}}\textbf{87.9}\textsubscript{0.7}\\ \color{blue}6.3\textsubscript{0.6}\end{tabular}     &
\begin{tabular}[c]{@{}l@{}}\textbf{93.4}\textsubscript{0.6}\\ \color{blue}\textbf{7.7}\textsubscript{0.6}\end{tabular} & 
\begin{tabular}[c]{@{}l@{}}48.3\textsubscript{1.9}\\ \color{blue}\textbf{18.7}\textsubscript{10.1}\end{tabular} &
\begin{tabular}[c]{@{}l@{}}77.1\textsubscript{2.0}\\ \color{blue}\textbf{14.2}\textsubscript{1.3}\end{tabular} &
\begin{tabular}[c]{@{}l@{}}74.3\textsubscript{0.8}\\ \color{blue}4.3\textsubscript{3.3}\end{tabular} & 
\begin{tabular}[c]{@{}l@{}}40.4\textsubscript{6.0}\\ \color{blue}\textbf{7.5}\textsubscript{3.2}\end{tabular} &  
\begin{tabular}[c]{@{}l@{}}\textbf{63.1}\textsubscript{14.0}\\ \color{blue}41.1\textsubscript{29.6}\end{tabular} \\ 
\bottomrule
\end{tabular}
\caption{Comparisons of DNIP results on varying optimization set sizes; min. optimization set size is 10 for all datasets except for DBpedia, which we use 15 to cover its 14 classes. For full optimization set sizes, AGNews: 9,500, DBpedia: 9,500, SST-5: 8,116, TREC: 5,179, RTE: 2,365, DDI: 9,500, PubMedQA: 950.}
\label{tab:varytrain_full}
\end{table*}

\end{document}